\documentclass{article}

 \usepackage[preprint]{neurips_2025}

\usepackage[utf8]{inputenc} % allow utf-8 input
\usepackage[T1]{fontenc}    % use 8-bit T1 fonts
\usepackage{hyperref}       % hyperlinks
\usepackage{url}            % simple URL typesetting
\usepackage{booktabs}       % professional-quality tables
\usepackage{amsfonts}       % blackboard math symbols
\usepackage{nicefrac}       % compact symbols for 1/2, etc.
\usepackage{microtype}      % microtypography
\usepackage{xcolor}         % colors
\usepackage[dvipsnames,svgnames]{xcolor}

\def\E{{\mathcal E}}
\def\C{{\mathcal C}}
\def\S{{\mathcal S}}

\usepackage{lineno}
\usepackage[most]{tcolorbox}
\usepackage{xcolor}
\usepackage{listings}
\usepackage{enumitem}
\usepackage{subcaption}
\usepackage{graphicx}
\definecolor{darkblue}{rgb}{0, 0, 0.5}
\hypersetup{colorlinks=true, citecolor=darkblue, linkcolor=darkblue, urlcolor=darkblue}
\definecolor{promptcolor}{RGB}{245, 245, 250} % Very light gray-blue

\newtcolorbox{promptbox}{
  colback=promptcolor,
  colframe=blue!50!black,
  boxrule=0.5pt,
  arc=2pt,
  left=4pt,
  right=4pt,
  top=4pt,
  bottom=4pt,
  width=\columnwidth, % << important change
  before skip=8pt,
  after skip=8pt,
}

\newtcolorbox{promptbox_whole}{
  colback=promptcolor,
  colframe=blue!50!black,
  boxrule=0.5pt,
  arc=2pt,
  left=4pt,
  right=4pt,
  top=4pt,
  bottom=4pt,
  width=\textwidth, % spans full page width
  before skip=8pt,
  after skip=8pt,
}

\usepackage{siunitx}
\newcommand{\prop}[1]{\num[round-mode=places, round-precision=4]{#1}}

\title{Reasoning with Preference Constraints: A Benchmark for Language Models in Many-to-One Matching Markets}

\author{%
  Marylou Fauchard \\
  Université de Montréal, Mila \\
  \texttt{marylou.fauchard@mila.quebec} \\
   \And
   Florian Carichon \\
   McGill, Mila \\
   \texttt{florian.carichon@mila.quebec}
   \And
   Margarida Carvalho \\
   Université de Montréal, Mila \\
   \texttt{carvalho@iro.umontreal.ca} \\
   \And 
   Golnoosh Farnadi \\
   McGill, Mila \\
   \texttt{farnadig@mila.quebec}
}

\begin{document}

\maketitle

\begin{abstract}
%Recent advances in reasoning with large language models (LLMs) have demonstrated strong performance on complex mathematical tasks, including combinatorial optimization. Techniques such as Chain-of-Thought and In-Context Learning have further enhanced this capability, making LLMs both powerful and accessible tools for a wide range of users, including non-experts. However, applying LLMs to matching problems, which require reasoning under preferential and structural constraints, remains underexplored. To address this gap, we introduce a novel benchmark of 369 instances of the College Admission Problem, a canonical example of a matching problem with preferences, to evaluate LLMs across key dimensions: feasibility, stability, and optimality. We employ this benchmark to assess the performance of several open-weight LLMs. Results reveal that while LLMs can satisfy certain constraints, they struggle to meet all evaluation criteria consistently, yet reasoning-augmented LLMs, like QwQ and GPT-oss, significantly outperform base ones, like the traditional Llama, Qwen or Mistral models, defined here as models used without any dedicated reasoning mechanisms.  Different LLMs reacted differently to the various prompting strategies tested which include Chain-of-Thought, In-Context Learning and role based prompting, with no prompt consistently offering the best performance. Performance from iterative prompting with auto-generated feedback is not monotonic; it can peak early and then significantly decline in later attempts.  
Recent advances in reasoning with large language models (LLMs) have demonstrated strong performance on complex mathematical tasks, including combinatorial optimization. Techniques such as Chain-of-Thought and In-Context Learning have further enhanced this capability, making LLMs both powerful and accessible tools for a wide range of users, including non-experts. However, applying LLMs to matching problems, which require reasoning under preferential and structural constraints, remains underexplored. To address this gap, we introduce a novel benchmark of 369 instances of the College Admission Problem, a canonical example of a matching problem with preferences, to evaluate LLMs across key dimensions: feasibility, stability, and optimality. We employ this benchmark to assess the performance of several open-weight LLMs. Our results first reveal that while LLMs can satisfy certain constraints, they struggle to meet all evaluation criteria consistently. They also show that reasoning LLMs, like QwQ and GPT-oss, significantly outperform traditional models such as Llama, Qwen or Mistral, defined here as models used without any dedicated reasoning mechanisms. Moreover, we observed that LLMs reacted differently to the various prompting strategies tested, which include Chain-of-Thought, In-Context Learning and role-based prompting, with no prompt consistently offering the best performance. Finally, we report the performances from iterative prompting with auto-generated feedback and show that they are not monotonic; they can peak early and then significantly decline in later attempts. Overall, this work offers a new perspective on model reasoning performance and the effectiveness of prompting strategies in combinatorial optimization problems with preferential constraints.

\end{abstract}

\section{Introduction} \label{intro}

Recent advancements in large language models (LLMs) have revealed emergent reasoning capabilities, enabling them to address increasingly complex problems. Most notably, LLMs have been applied to solve mathematical problems that require step-by-step analytical reasoning \citep{ahn2024large}. They have also been used in complex game-theoretic scenarios with interactions and strategic decision-making \citep{sun2025game} and combinatorial optimization tasks \citep{yang2023large,wang2023can}, proving that LLMs can extend their reasoning abilities to various kind of complex real-world scenarios. To rigorously assess the effectiveness of LLM reasoning and its associated methods, a range of benchmarks and evaluation metrics have been developed. These benchmarks can be categorized into (1) objective, ground-truth-based benchmarks such as MATH  \citep{hendrycks2021measuring} or SCIBENCH \citep{wang2023scibench} and (2) subjective, preference-based evaluations such as Chatbot Arena \citep{chiang2024chatbot}. Although these benchmarks are valuable for understanding LLMs' capacity to solve these theoretical problems, they are still limited due to static question sets, the risk of data contamination, a lack of real-world complexity, high variance, or prioritizing form over reasoning quality \citep{chen2025heurigym}. Combinatorial optimization benchmarks such as HeuriGym \citep{chen2025heurigym}, CO-Bench \citep{sun2025co}, or FrontierCO \citep{feng2025comprehensive} provide ideal evaluation settings as they involve multi-step and iterative reasoning over the large solution spaces of NP-hard problems, resist memorization well, and offer robust quantitative metrics \citep{sartori2025combinatorial,chen2025heurigym}. While these benchmarks show that LLMs hold promise in solving NP-hard problems with cost-minimization objectives, they overlook a critical class of objective combinatorial problems involving preference-based constraints, where stability among multiple agents is the primary measure of success.

Many-to-one matching problems, which are solvable in polynomial time and involve satisfying preferences and capacity constraints, 
provide a natural benchmark with key properties for testing this aspect of LLMs' reasoning abilities. First, non-experts often address this optimization problem in several contexts, such as College Admission \citep{sartori2025combinatorial}, Hospital-Resident matching \citep{roth1984}, or School Choice \citep{scp2003}. These problems' model has been widely adopted by institutions to represent participants' preferences, such as in the US \citep{nyc2005,boston2005,roth1984}, Hungary \citep{hungary2008}, and Germany \citep{germany2007}.
Secondly, the problem involves providing a matching, notion formally defined in Section~\ref{problem_statement}, that satisfies desirable properties regarding the students' and colleges' preferences, introducing an additional layer of reasoning complexity. Specifically, the objective is to find a matching that minimizes the sum of ranks for students, such that no college exceeds its capacity and that no student-college pair prefers to be matched to each other over their assigned outcomes. Such a solution is considered student-optimal, feasible, and stable~\citep{gale1962college}. %These characteristics make it an ideal testbed for assessing whether LLMs can solve complex reasoning problems by respecting user constraints and objectives. 
In the remainder of this article, we will specifically address the College Admission problem.
Our contributions are the following: 
\begin{itemize}
\item We introduce a new benchmark dataset with objective metrics specifically designed for preferential constraint problems, namely the College Admission problem.
\item We conduct an empirical evaluation of several open-source LLMs, including models from Llama, Qwen, Mistral and GPT, under various prompting strategy which vary in training paradigms and model sizes.
\item By analyzing the feasibility, stability, and optimality of the solutions generated by LLMs, we investigate how various dimensions of problem complexity, given by the number of students and their preferences, influence their reasoning capabilities.
\item We demonstrate how several prompting approaches, such as Chain-of-Thoughts (CoT), In-Context Learning (ICL), or iterative prompting, could influence the reasoning of LLMs in solving matching problems.
\end{itemize}

\section{Related Word}
\label{related_work}
This work builds on two key areas of related research: two-sided many-to-one matching markets and recent developments and benchmarks for LLMs' reasoning.

\subsection{Two-Sided Many-to-One Matching Markets}
Matching markets are mechanisms designed to pair agents on two sides of a market based on their preferences. A foundational model is the one-to-one matching problem, famously formalized by Gale and Shapley in the context of the Stable Marriage Problem \citep{gale1962college}, where each participant on one side is matched with exactly one on the other. Many real-world applications, however, fall under the many-to-one setting.
This setting mainly allows an unequal number of agents on both sides, with different capacities making the problem more challenging. 
%Some of the most well-known problems in many-to-one matching markets include the Hospital-Resident matching problem \citep{roth1984}, College Admissions \citep{gale1962college}, and School Choice programs \citep{scp2003}. 
The Deferred Acceptance (DA) algorithm, introduced by Gale and Shapley \citep{gale1962college}, has become a foundational tool for solving both one-to-one and, its generalization, many-to-one matching problems. The resulting matching is guaranteed to be feasible, stable, and student-optimal, notions that will be formally defined in Section \ref{problem_statement}. This algorithm has been implemented in several real-world settings \citep{nyc2005,boston2005,roth1984,hungary2008}. 
They demonstrate that a diversity of institutional constraints, such as stability or equitable access, exists in the College Admission problem, and more generally in many-to-one problems. They emphasize the dimensions that can be evaluated in LLMs complex reasoning.
Recent work has focused on extensions and generalizations of these matching problems by introducing additional layers of complexity, such as allowing ties in preferences and enforcing common quotas \citep{agoston2022college},  considering flexible capacities \citep{bobbio2022,bobbio2023,afacan2024capacity} or contingent priorities~\citep{rios2024stable}. While existing datasets, such as the Chilean school choice dataset~\citep{10.1145/3328526.3329580}, provide valuable real-world data, they do not offer sufficient control over problem instance parameters to effectively evaluate an LLM's reasoning and understanding.

\subsection{LLMs Reasoning Benchmarks}
LLMs have recently demonstrated strong capabilities to tackle complex reasoning tasks in various fields \citep{parashar2025inference}. In the context of mathematical reasoning, LLMs are primarily evaluated for their capabilities in logical analysis, deductive reasoning, and arithmetic operations \citep{yuan2023well}. Standard benchmarks such as MATH \citep{hendrycks2021measuring}, Olympiad-Bench \citep{he2024olympiad}, or MathBench \citep{liu2024mathbench} have become central in evaluating reasoning capabilities of LLMs. Models have achieved high performances on traditional benchmarks, raising concerns about memorization instead of true reasoning capabilities with multi-step deduction, calling for a new perturbed dataset \citep{huang2025mathperturb}. 
Alongside these benchmarks, combinatorial optimization problems offer a rich ground for evaluating complex reasoning. They exhibit specific characteristics: they often have numerous potential solutions that are not always directly comparable, involve different constraints on variables, and the optimization process can encounter multiple local optima \citep{sartori2025combinatorial}. Within this framework, LLMs have been tested for heuristic generation \citep{ye2024reevo}, algorithm design \citep{sartori2025combinatorial}, and direct problem-solving via prompt-based techniques \citep{guo2023towards,yang2023large}. The first existing benchmarks focus on the classical and potentially overused Traveling Salesman Problem (TSP) \citep{sartori2025combinatorial,ye2024reevo,yang2023large}. Other benchmarks mainly assess reasoning through real-world scenarios based on graph and set optimization problems \citep{wang2023can, feng2025comprehensive}, pairing and scheduling problems \citep{ chen2025heurigym, sun2025co}, or machine learning related problems such as logistic regression or grid search optimization \citep{guo2023towards,yang2023large}. Most of these benchmarks predominantly target well-known NP-hard combinatorial problems centered on cost minimization heuristics. Therefore, they neglect structured tasks, such as problems with preference-based constraints and stability requirements. A first step toward addressing constraint-driven problems has been explored with the Transportation and Assignment Problems \citep{khan2024capability}.  However, although the underlying mathematical formulations support many-to-one matchings, the prompt design appears to restrict the problem to one-to-one instances. Recent work has also studied LLMs understanding of one-on-one matching market with rigorous evaluation \citep{matchingLLM2025}, but the many-to-one market offers a better testbed for evaluating LLM reasoning with more complexity due to potentially unmatched agents and respecting complex capacity constraints.  
This article introduces a novel benchmark for evaluating LLMs on the College Admission problem. By implementing and comparing several prompting strategies, our benchmark not only exposes some current limitations of LLMs in tackling complex combinatorial optimization tasks but also reveals interesting dynamics for evaluating reasoning complexity and model understanding in structured decision-making contexts.

\section{Methodology}
\label{methodology}

\subsection{Problem Statement} \label{problem_statement}

%In this section, we introduce the formal notation for the College Admission problem to clarify terminology and support the definition of our evaluation metrics.
In the College Admission problem, we have a set of students $\mathcal{S}$ and a set of colleges $\mathcal{C}$. Each student in $s \in \mathcal{S}$ has a strict preference order $\succ_s$ over $\mathcal{C}\cup\{\emptyset\}$. We write $c_i \succ_s c_j$ for $c_i, c_j \in \mathcal{C}$ if student $s$ prefers  college $c_i$ over $c_j$. If student $s$ prefers to be unassigned than being matched with a college $c \in \mathcal{C}$, we write $\emptyset \succ_s c$. Therefore, the student's preference list results in a ranking of the colleges. We denote by $rank_s(c)$ the rank of a college $c \in \mathcal{C}$ in the preference list of student $s$, e.g., the most preferred college for student $s$ has rank 1. Similarly, each college $c \in \mathcal{C}$ has a strict order $\succ_c$ over $\mathcal{S} \cup \{\emptyset\}$. Moreover, a college $c$ has a capacity $q_c \in \mathbb{Z}_+$. We denote \(\mathcal{E} \subseteq \mathcal{S} \times (\mathcal{C} \cup \{\emptyset\})\) as the set of acceptable pairs, meaning that $(s,c) \in \mathcal{E}$ if $c \succ_s \emptyset$ and $s \succ_c \emptyset$. 

An assignment is any subset \(\mathcal{M} \subseteq \mathcal{E}\), interpreted as any set of student–college pairs. An assignment does not necessarily satisfy capacity constraints. An assignment $\mathcal{M}$ is \emph{assignment stable} if for no pair $(s,c)\in \E \setminus \mathcal{M}$, (i) student $s$ prefers $c$ over their current matching, i.e., $c\succ_s \mathcal{M}(s)$ where $\mathcal{M}(s)$ is either the school assigned to student $s$ or $\emptyset$ when the student is unmatched, (ii) school $c$ is under-subscribed, i.e., $|\mathcal{M}(c)|<q_c$ where $\mathcal{M}(c)$ is the set of students assigned to $c$, or prefers student $s$ over some student $s'\in \mathcal{M}(c)$, i.e., $s\succ_c s'$. A pair that would respect either of the previous conditions is called a blocking pair and prevents stability.

A set $\mu \subseteq \E$ is called a \emph{feasible matching} when (i) each student $s\in \S$ is matched to at most one school, i.e.,  $|\{c \in \C: (s,c) \in \mu \}| \leq 1$, and (ii) each school $c \in \C$ is matched to no more students than its capacity, i.e., $|\mu(c)=\{s \in \S: (s,c) \in \mu\}|\leq q_c$. For clarity, we will refer to feasible matchings as "matchings" in the remainder of the paper. A matching $\mu$ is \emph{matching stable} if for no pair $(s,c)\in \E \setminus \mu$, (i) student $s$ prefers $c$ over their current matching, i.e., $c\succ_s \mu(s)$ where $\mu(s)$ is either the school assigned to student $s$ or $\emptyset$ when the student is unmatched, (ii) school $c$ is under-subscribed, i.e., $|\mu(c)|<q_c$ where $\mu(c)$ is the set of students assigned to $c$, or prefers student $s$ over some student $s'\in \mu(c)$, i.e., $s\succ_c s'$. By default, throughout the paper, the term stability will refer to matching stability. Finally, a stable matching $\mu$ is \emph{student-optimal} when no student can get a better assignment in any other stable matching. It is known that a student-optimal stable matching $\mu$ coincides with the stable matching minimizing the sum of the ranks of the students, i.e., $\sum_{(s,c) \in \mu} rank_s(c)$. With these formalizations, we define our College Admission problem as computing the student-optimal stable matching.

\subsection{Benchmark}\label{benchmark_details}

We introduce a new benchmark dataset explicitly designed for controlled experimentation. Compared to the Chilean dataset \citep{10.1145/3328526.3329580}, which focuses on large-scale problems, our benchmark provides a controlled framework over instance parameters, enabling systematic evaluation on well-defined problem settings.
This controlled environment enables detailed analysis of model reasoning and decision-making processes. Additionally, the code to generate the benchmarks data\footnote{\label{footnote}The link to the GitHub page will be added upon acceptance.} allows users to scale instance complexity by adjusting the number of students, colleges, and other parameters to accommodate broader generalization experiments.

In this article, we generate synthetic instances of the College Admission problem, varying the number of students across four levels: 5, 10, 15, and 20. While the number of students is limited in this setting, our benchmark is designed to be scalable, allowing easy extension to real-world sizes while remaining compatible with other experimental parameters.
Each configuration is tested under three types of student preference structures: 
\textbf{Complete Preferences} where each student ranks all available colleges; \textbf{Incomplete Preferences} where each student ranks a fixed number of colleges, strictly fewer than the total available; and \textbf{Flexible Preferences} where the length of each student's preference list is drawn randomly between a predefined minimum and the total number of colleges. Since college preferences do not impact the complexity of the DA algorithm, they will always remain complete. Moreover, we vary the total capacity of the system across three settings: \textbf{Under-capacity} with a capacity of 80\% of the number of students, forcing some students to be unassigned; \textbf{Exact-capacity} where the total capacity equals the number of students, and \textbf{Over-capacity}  
where total capacity exceeds the number of seats by 10.
%Under fixed preferences, more capacity should not change the outcome if the model is reasoning correctly. 
We want to test this configuration since adding capacity to certain colleges should not impact the solution, assuming that preferences remain unchanged, if the model reasons correctly. 
Colleges are assigned random capacities, ensuring that each college has at least one available seat. We also explore different student-to-college ratios (1:1, 2:1, 3:1, 4:1), with some configurations omitted when the total capacity falls below the number of colleges. For each combination of parameters, we generate three random seeds while creating the preferences to account for variability. In total, our benchmark comprises 369 instances, with some configurations omitted due to infeasibility. A more detailed breakdown of the number of instances is included in Section~\ref{instance_prompt}.

\subsection{Prompts}\label{prompt}
\textit{\textbf{Prompt strategy}}
%Prompts structure and formatting
with distinct formatting significantly influence LLM performance \citep{mao2025,liu2025beyond}. To rigorously evaluate LLM performance on this reasoning task,  our 369 instances are augmented with various prompt variations. We evaluate the effect of advanced prompting strategies for LLMs' reasoning, such as CoT prompting \citep{chen2025towards}, ICL \citep{dong2022survey}, and role-based prompting \citep{sartori2025combinatorial}.
More specifically, the prompt containing only the essential information is referred to as the \textbf{Basic} prompt. The \textbf{Role} prompt follows established prompt templates from prior work \citep{sartori2025combinatorial}. This component is typically placed at the beginning of the prompt to instruct the model on the intended perspective or behavior \citep{mao2025}. Two of the prompting strategies fall under the category of ICL. The example is included immediately following the output format component, as commonly adopted in prior work \citep{mao2025}. The first one, simply named the \textbf{ICL}, consists of adding a simplified problem instance along with its corresponding student-optimal matching.
To maintain the problem's core structure while adopting a simple example, we use a five-student instance with consistent student-to-school ratios, capacity types, and preference types across different random seeds. The other strategy, referred to as \textbf{ICL with steps} in the remainder of the paper, builds upon the findings of \cite{wei2022chain}, which suggest that incorporating intermediate reasoning steps into examples can enhance model performance. This strategy is implemented using a fixed example with five students, complete preference lists, and exact capacity settings, where intermediate steps correspond to those of the DA algorithm. The last four prompting strategies are variations of CoT prompting, which we categorize as \textbf{CoT unsupervised}, \textbf{CoT text}, \textbf{CoT pseudocode}, and \textbf{CoT Python}. The CoT unsupervised represents the 
traditional unguided \textit{Think step-by-step} instruction \citep{zhang2025prompt}.
In contrast, the CoT text variant follows a supervised template that includes an explanation of the solution process \citep{zhang2025prompt}. The step-by-step explanation is based on a \emph{natural language} description of the DA algorithm introduced in \cite{aziz}. Incorporating pseudocode instructions can also improve performance in graph reasoning tasks \citep{skianis2024}. To investigate whether similar benefits extend to matching problems, we include a pseudocode variation with a description guiding the step-by-step process. Finally, CoT prompting with Python-based, self-describing programs enhances performance on mathematical reasoning tasks \citep{jie2023design}. Accordingly, our CoT Python template includes an implementation of the DA algorithm written in Python.

\textit{\textbf{Iterative Prompting}} is a multi-step process that iteratively tries to improve LLMs' output by incorporating feedback on the previous failed attempts \citep{madaan2023selfrefineiterative}. Prior works have shown promising results from this method that requires no additional training \citep{ho2025selfcritiqueguidedcuriosityrefinementenhancing,yang2023large}. 
More specifically, starting from the original prompt, this technique consists of adding the last model response and feedback to the preceding prompt to improve the following generated answer, up to a predefined maximum of N attempts \citep{krishna2024understandingeffectsiterativeprompting}.
While much of the work focuses on AI-generated feedback, we use external feedback, which is non-AI-generated, as it is more reliable \citep{stechly2023gpt4doesntknowits}. In our setting, we test iterative prompting, with up to 5 iterations, adding feedback based on whether the previous output verifies our four metrics.

More details, including the structure of each prompt are available in Appendix~\ref{instance_prompt}.

\section{Empirical Evaluation}
\label{evaluation}
\begin{figure*}[t!]
     \centering
     \includegraphics[width=0.7\textwidth]{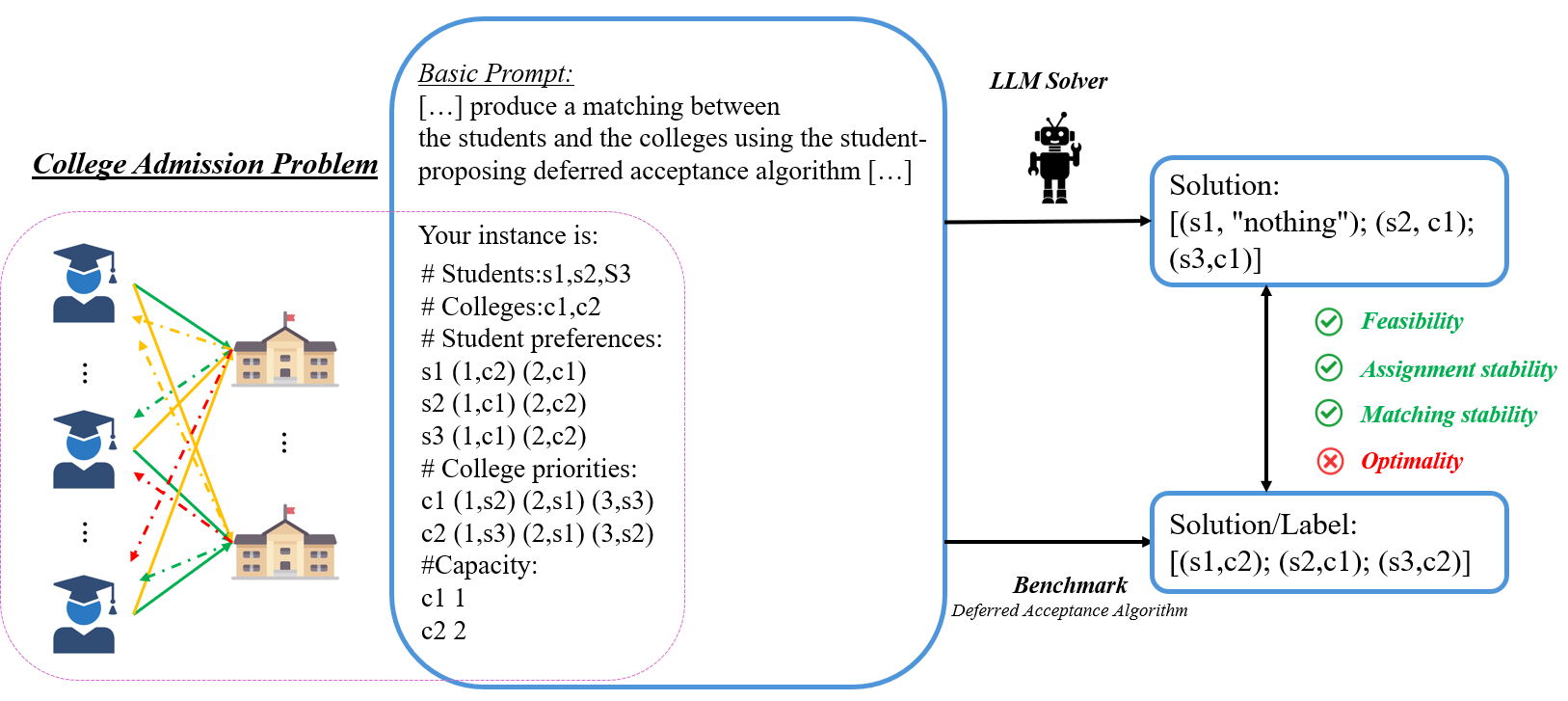}
     \caption{Experimental protocol for generating solutions to the College Admission Problem. We fed the LLM one instance associated with one prompt strategy. We compare the solution to the one of the DA algorithm to measure the feasibility, stability and optimality of the LLM-generated answer.}
     \label{fig:protocol}
 \end{figure*}
 
We evaluate performance across the following models: (1) \textbf{Llama 3 8B} \citep{metaLlama} (2) \textbf{Llama 70B} \citep{metaLlama} (3) \textbf{Mistral 7B} \citep{jiang2023mistral}, (4) \textbf{Qwen2 7B} \citep{qwen2}, (4) \textbf{QwQ 32B} \citep{qwq32b}, and (5) \textbf{GPT-oss 120B} \citep{agarwal2025gpt}. We have selected these open-weight models for transparency and reproducibility purposes, ensuring that we evaluate different reasoning abilities on complex problems. While the first four models are said to be able to solve complex reasoning problems, the latter have been trained for thinking and reasoning using reinforcement learning and supervised fine-tuning. Therefore, for the remainder of the paper, we will refer to the first four models as the base models and the last two as the reasoning models. We specifically aim to assess their reasoning capabilities in solving our College Admission problem and analyze the impact of model size, model training, and problem complexity on solution quality. Figure~\ref{fig:protocol} introduces our experimental protocol for generating solutions using various LLM models.

To evaluate the quality of the generated matchings, we examine four key properties: feasibility, assignment stability, matching stability, and student optimality, which are formally introduced in Section~\ref{problem_statement}. Since feasibility only consists of respecting the capacities explicitly mentioned in the instance and assignment stability requires understanding the concept of blocking pairs and applying it to the given preferences, we consider that the various metrics introduced: (1a) feasibility, (1b) assignment stability, (2) matching stability, and (3) student-optimality form an ascending hierarchy of difficulty, as each level presupposes the satisfaction of the previous one.
%While feasibility only consists of respecting the capacities explicitly mentioned in the instance, assignment stability requires understanding the concept of blocking pairs and applying it to the given preferences. We thus consider feasibility as the easiest metric and assignment stability as harder as it is more implicit. Also, between feasibility, matching stability and student-optimality, the metrics are  increasingly difficult because they require the success of previous ones.
A response is considered valid if it adheres to the expected output format of the DA algorithm, allowing for minor acceptable variations such as using parentheses instead of brackets or different separators between pairs.
Table~\ref{count_table} presents the percentage of valid outputs. Considering only the valid outputs, we compute the proportion of matchings that satisfy each metric.
The dataset and the code are available on our GitHub page in footnote~\ref{footnote}. 

Our evaluation is guided by the following research questions:

\paragraph{RQ1: How well do LLM-generated matchings satisfy feasibility, stability, and student-optimality?}
\begin{figure*}[t]
  \centering
  \begin{subfigure}{0.45\textwidth}
    \includegraphics[width=\linewidth]{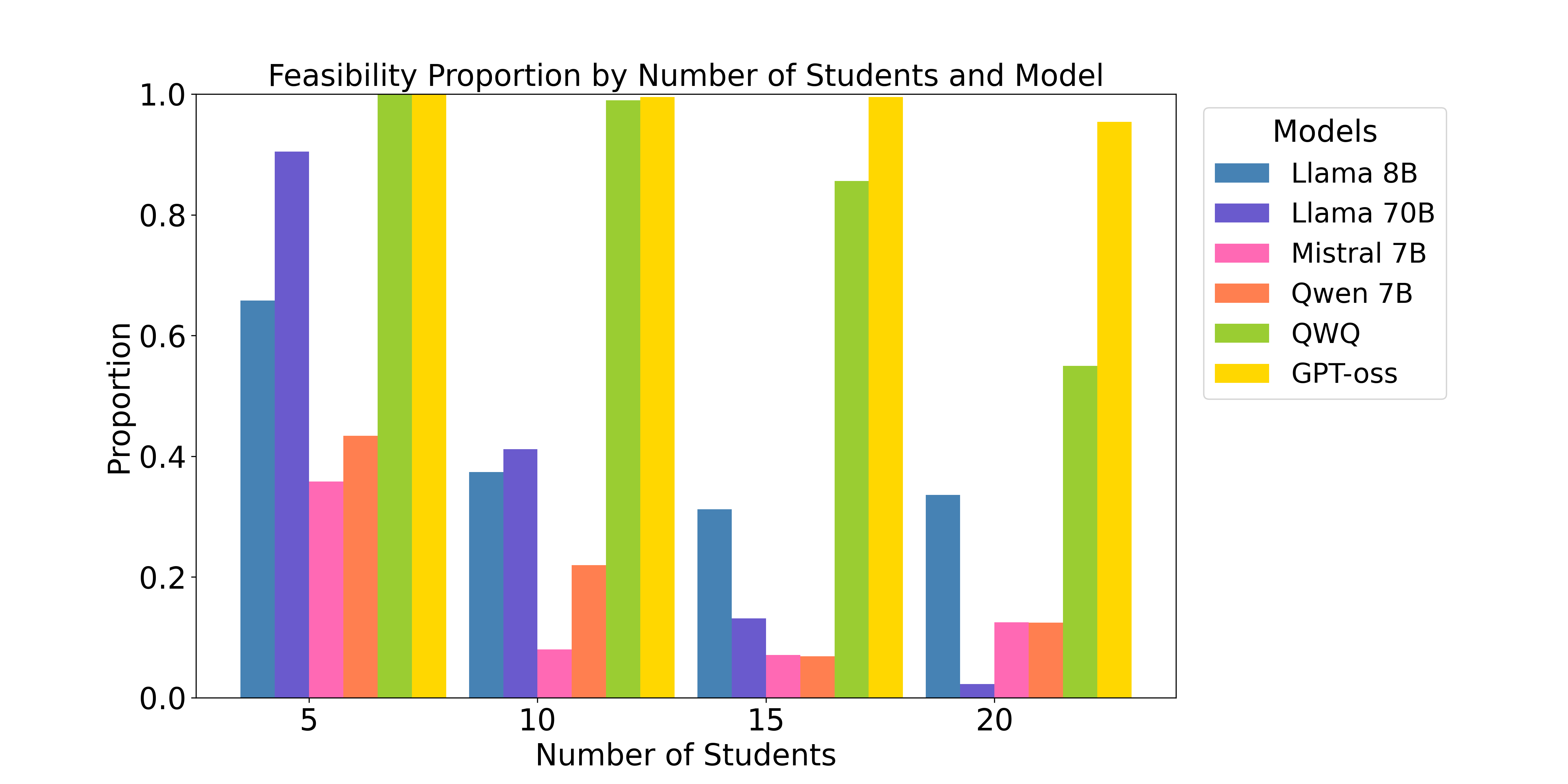}
    \caption{Feasibility}
    \label{fig:feasible_students}
  \end{subfigure}
  \hfill
  \begin{subfigure}{0.45\textwidth}
    \includegraphics[width=\linewidth]{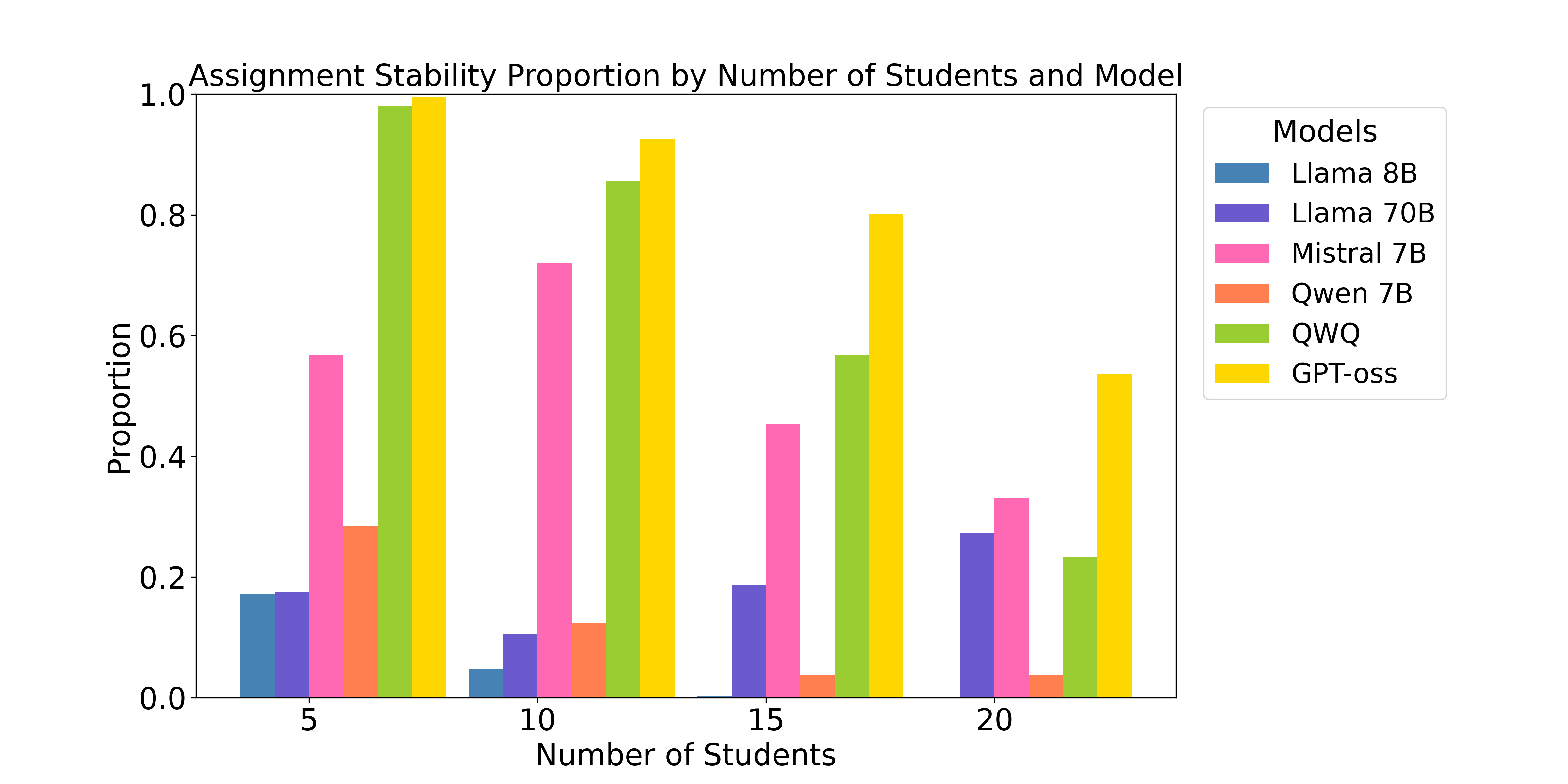}
    \caption{Assignment Stability}
    \label{fig:assignment_students}
  \end{subfigure}
  
  \vspace{0.5em}
  
  \begin{subfigure}{0.45\textwidth}
    \includegraphics[width=\linewidth]{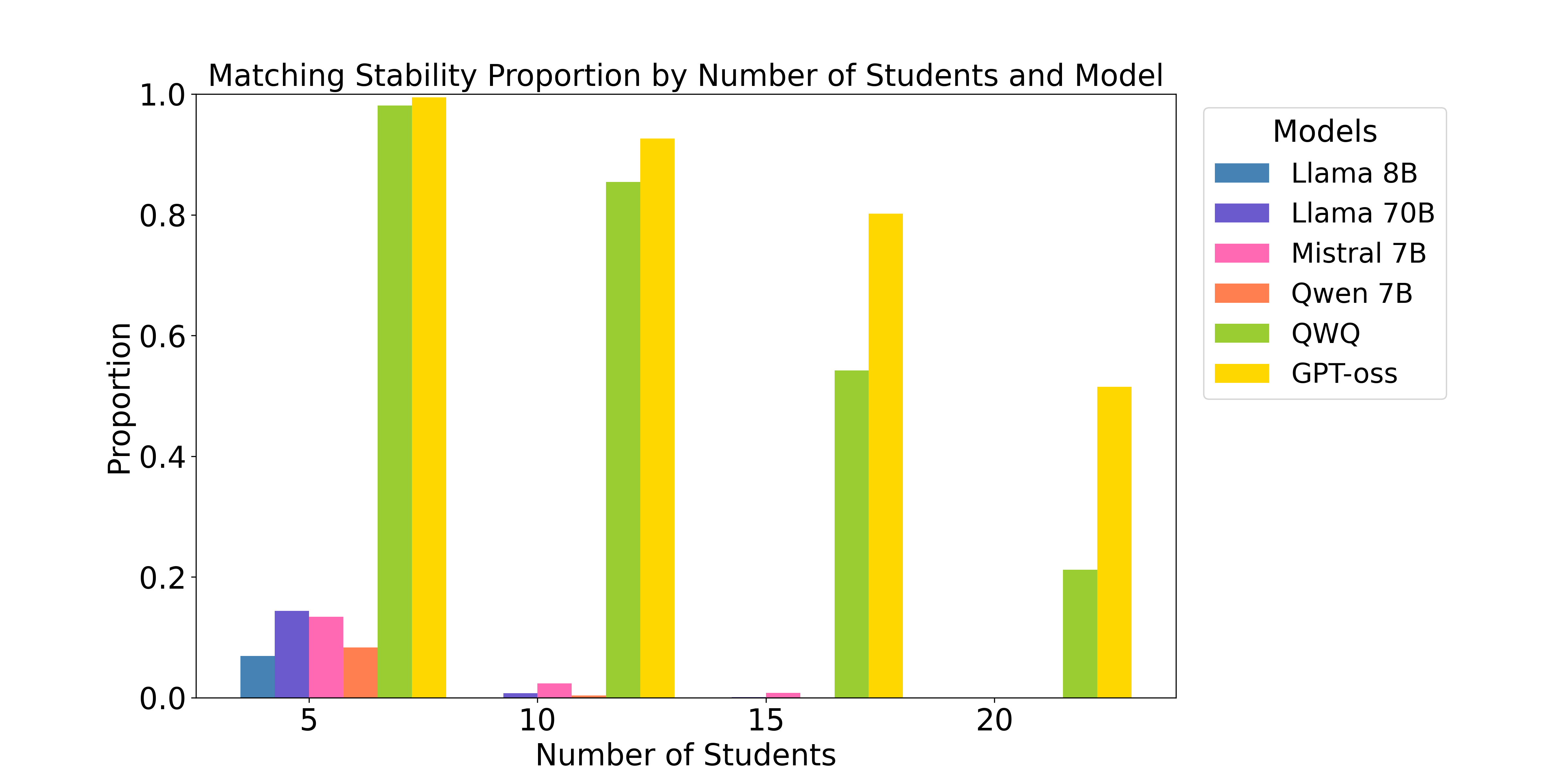}
    \caption{Matching Stability}
    \label{fig:matching_students}
  \end{subfigure}
  \hfill
  \begin{subfigure}{0.45\textwidth}
    \includegraphics[width=\linewidth]{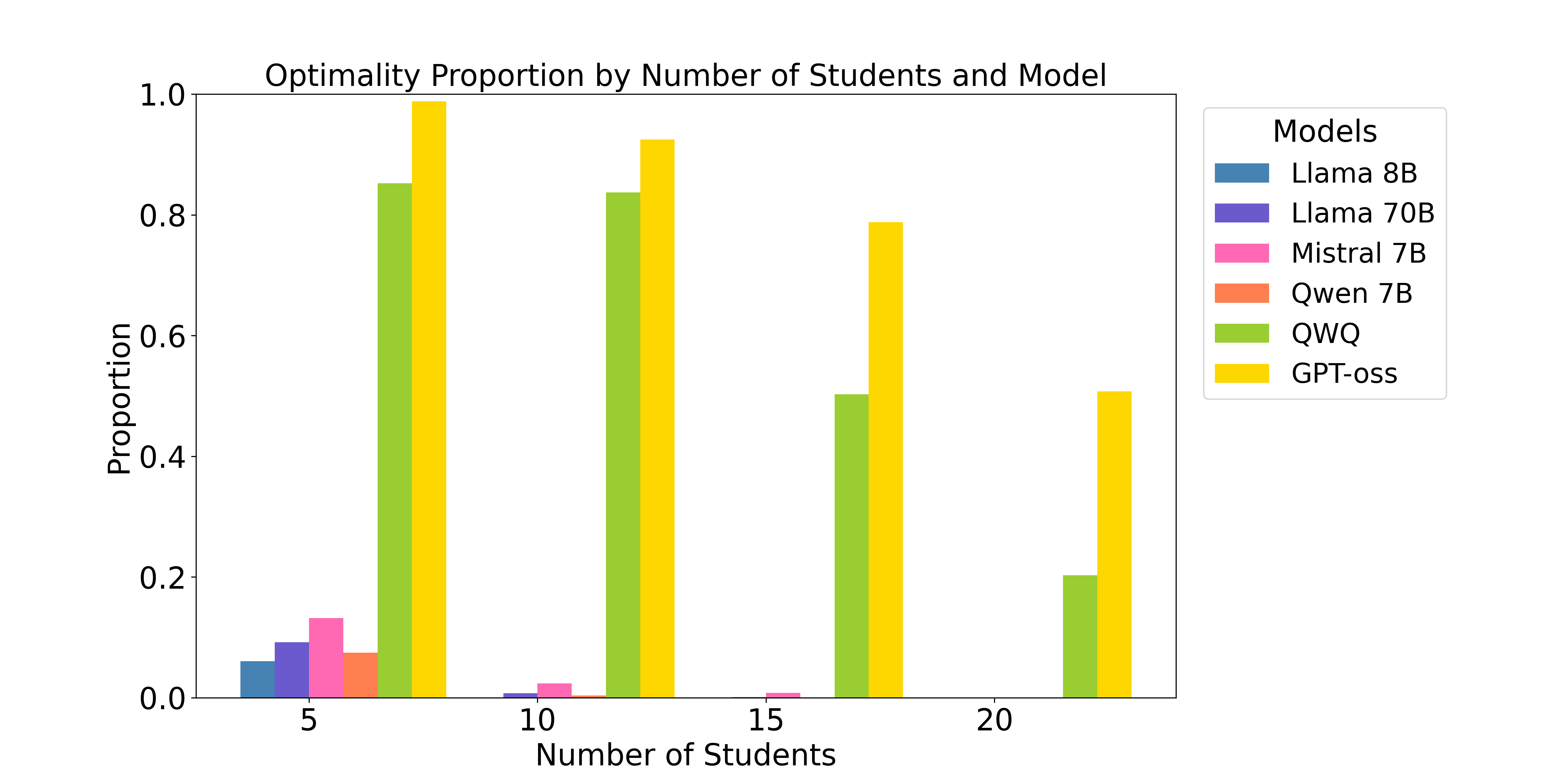}
    \caption{Optimality}
    \label{fig:optimal_students}
  \end{subfigure}

  \caption{Proportion metrics by number of students and models. Results are aggregated over all prompts.}
  \label{fig:students_metrics_model}
\end{figure*}

\begin{figure*}[]
  \centering

  \begin{subfigure}{0.45\textwidth}
    \includegraphics[width=\linewidth]{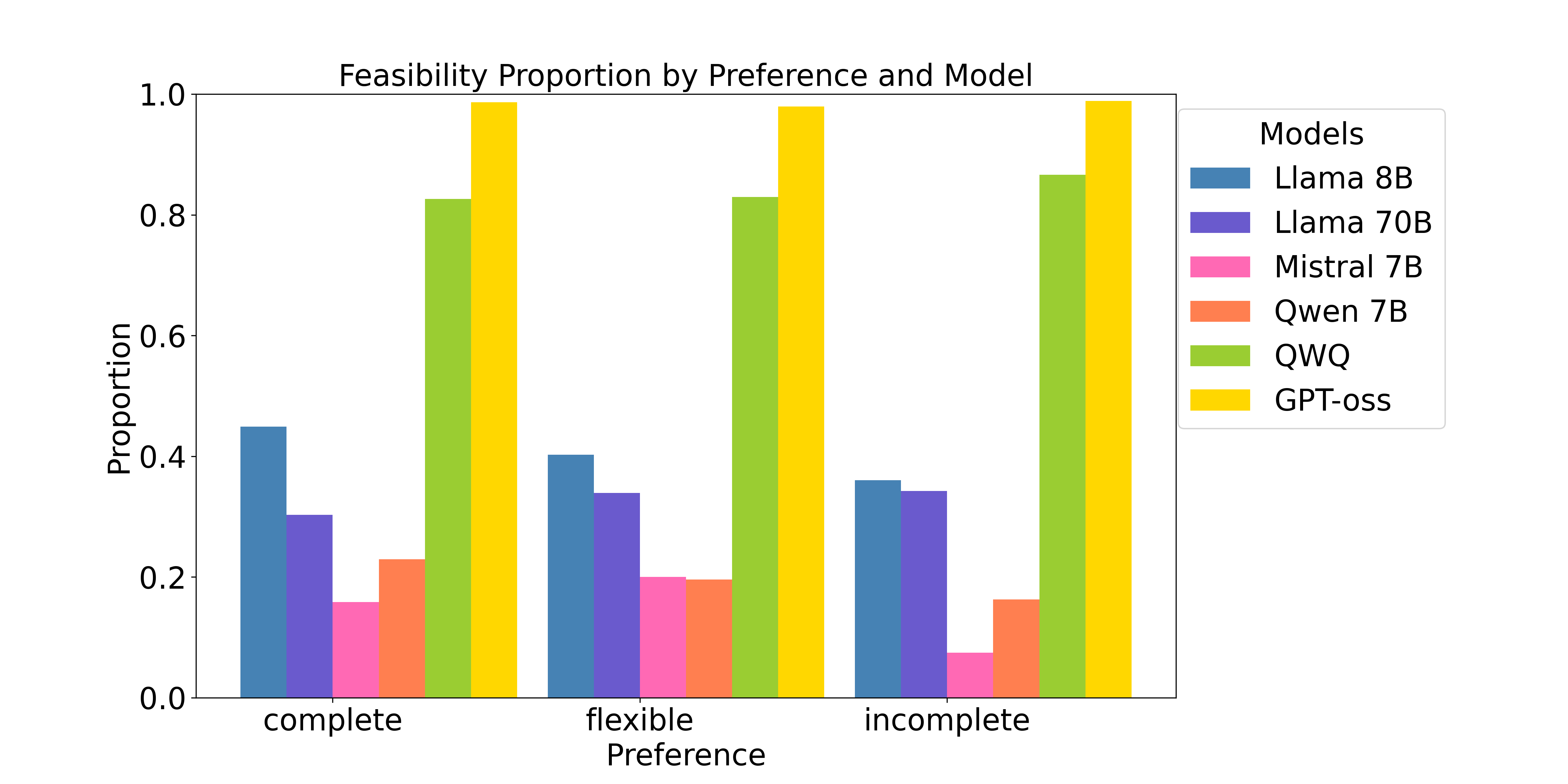}
    \caption{Feasibility}
    \label{fig:feasible_pref}
  \end{subfigure}
  \hfill
  \begin{subfigure}{0.45\textwidth}
    \includegraphics[width=\linewidth]{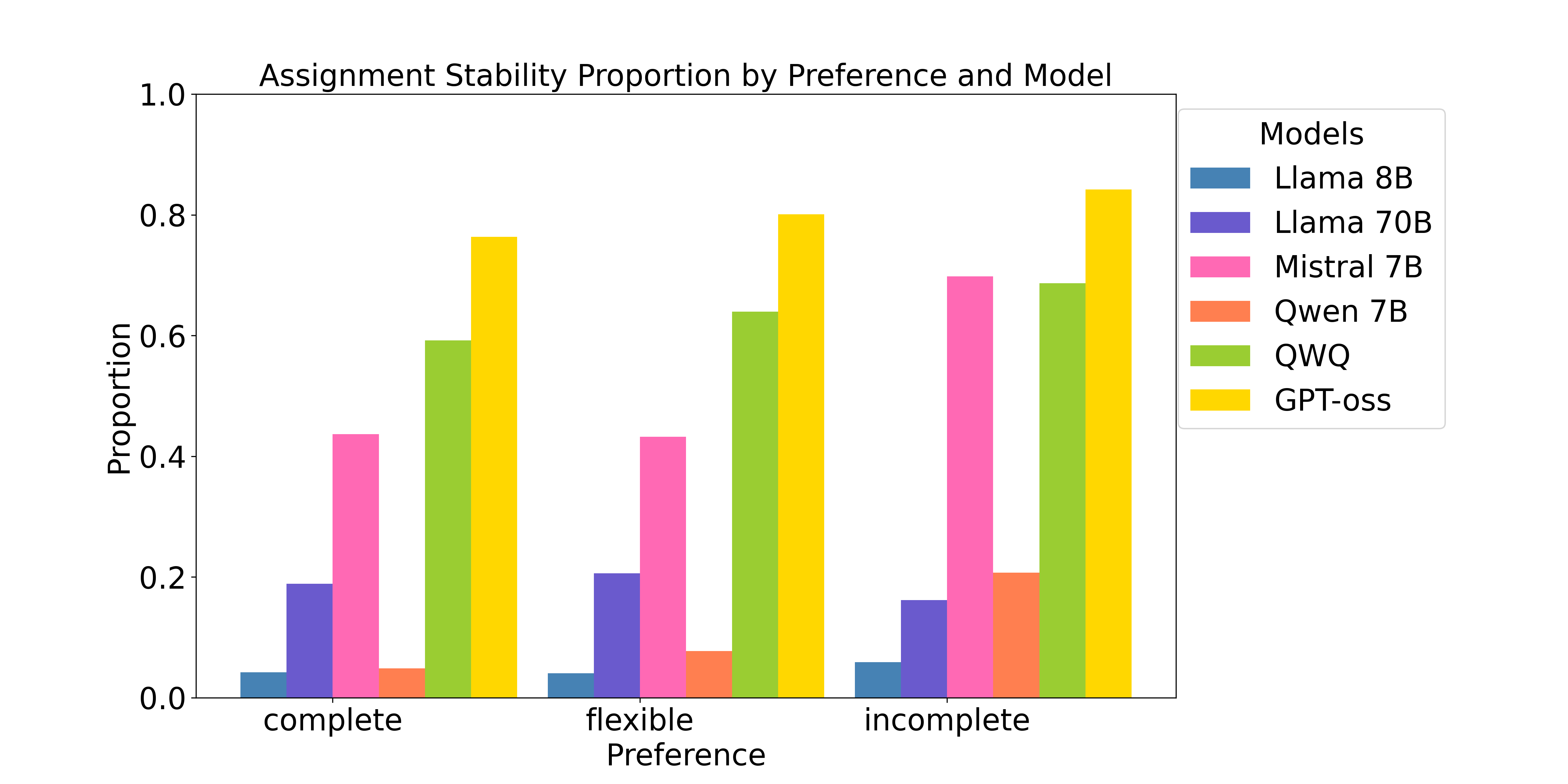}
    \caption{Assignment Stability}
    \label{fig:assignment_pref}
  \end{subfigure}

  \begin{subfigure}{0.45\textwidth}
    \includegraphics[width=\linewidth]{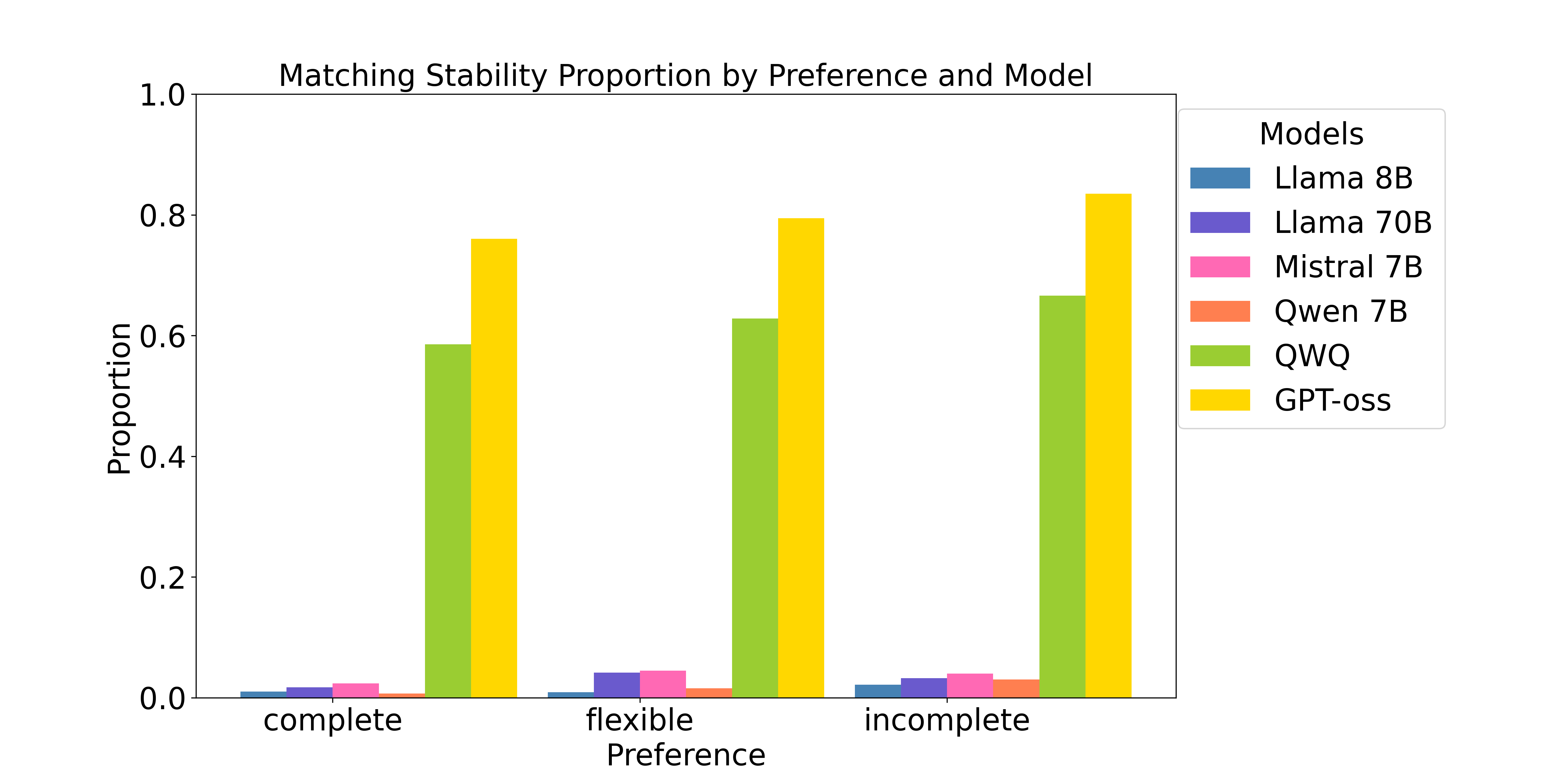}
    \caption{Matching Stability}
    \label{fig:matching_pref}
  \end{subfigure}
  \hfill
  \begin{subfigure}{0.45\textwidth}
    \includegraphics[width=\linewidth]{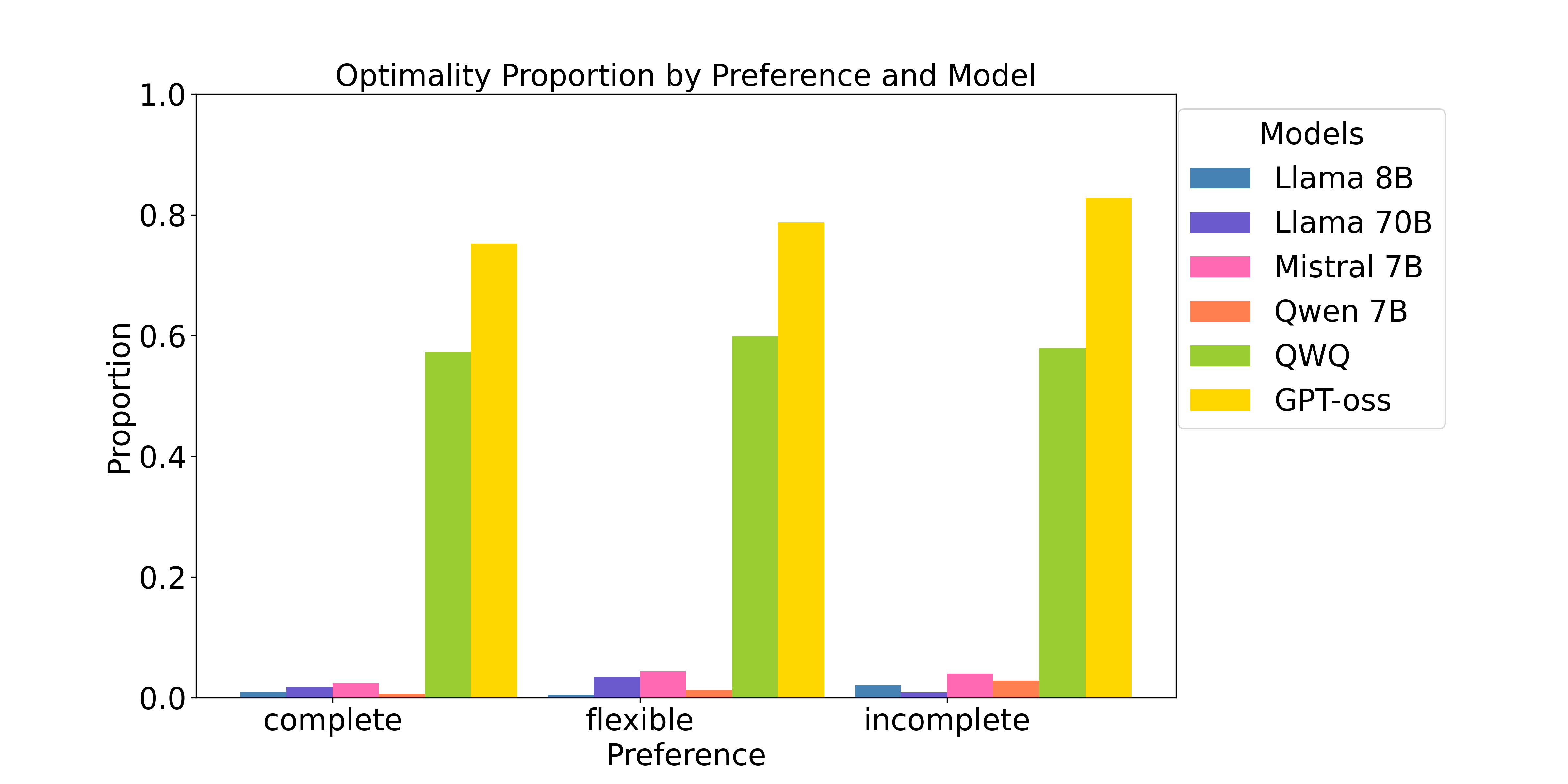}
    \caption{Optimality}
    \label{fig:optimal_pref}
  \end{subfigure}

  \caption{Normalized performance by preference for all models.}
  \label{fig:pref_model}
\end{figure*}

Figure~\ref{fig:students_metrics_model} presents the proportion of matchings, for each model, that satisfy each metric: feasibility (Figure~\ref{fig:feasible_students}), assignment stability (Figure~\ref{fig:assignment_students}), matching stability (Figure~\ref{fig:matching_students}), and optimality (Figure~\ref{fig:optimal_students}) based on the number of students. 
%Aggregated metrics by prompts can be found in Appendix~\ref{tables_graphs}, along with detailed results by model and prompt strategy in Table~\ref{overall_prompt_model}.
The results show that the matching problem is a challenging task, where base LLMs struggle even for the most explicit metric. The reasoning LLMs largely outperform the base ones on every metric and across instance sizes, highlighting the need for specialized training in reasoning. For all LLMs, performance decreases as instance size increases; however, advanced reasoning models exhibit a less abrupt drop in feasibility, indicating they can maintain track of such a simple and explicit metric. For stability and optimality, which requires understanding and preventing the notion of blocking pairs, the task is more challenging even for reasoning models. While the DA algorithm's complexity depends on the number of students, it also depends on the length of the students' preference lists. 
Figure~\ref{fig:pref_model} displays the metric distributions across different preference types and models. Figure~\ref{fig:pref_metrics_prompt} in the Appendix~\ref{tables_graphs} presents the metrics for every prompt strategies instead of models, where the following conclusions can still be drawn.
Complete preferences increase the algorithmic complexity by increasing the number of iterations required for an algorithm to arrive at the optimal solution. In contrast, incomplete preferences present more reasoning complexity, as they break completeness, creating more invalid pairs and requiring the algorithm to reason about acceptability and the presence of unmatched agents. In other words, incomplete preferences inherently increase the number of invalid pairs (i.e., the agent would rather remain unmatched than be assigned to a less preferred option). A matching that contains one invalid pair will always be unstable. However, the results suggest that the models do not struggle with this aspect. Performance remains almost unchanged with varying preference types, unlike the number of students, with a slight advantage for incomplete preferences.

\paragraph{RQ2: How does the prompting strategy and model choice affect the quality of the generated matchings? }
We previously mentioned that reasoning LLMs' performance dominates that of the base LLMs, especially when dealing with more complex instances and metrics. Although this shows that specialized training for reasoning is needed to solve even a polynomial algorithm,  %some models can grasp difficult concepts without this reasoning mode. 
Figure~\ref{fig:assignment_students} and Figure~\ref{fig:assignment_pref} reveal that Mistral can better apprehend the concept of stability with similar performances as the reasoning LLMs, but failed to perform well on the subsequent metrics because it can not satisfy feasibility as well. 

\begin{figure*}[t]
  \centering

  \begin{subfigure}{0.45\textwidth}
    \includegraphics[width=\linewidth]{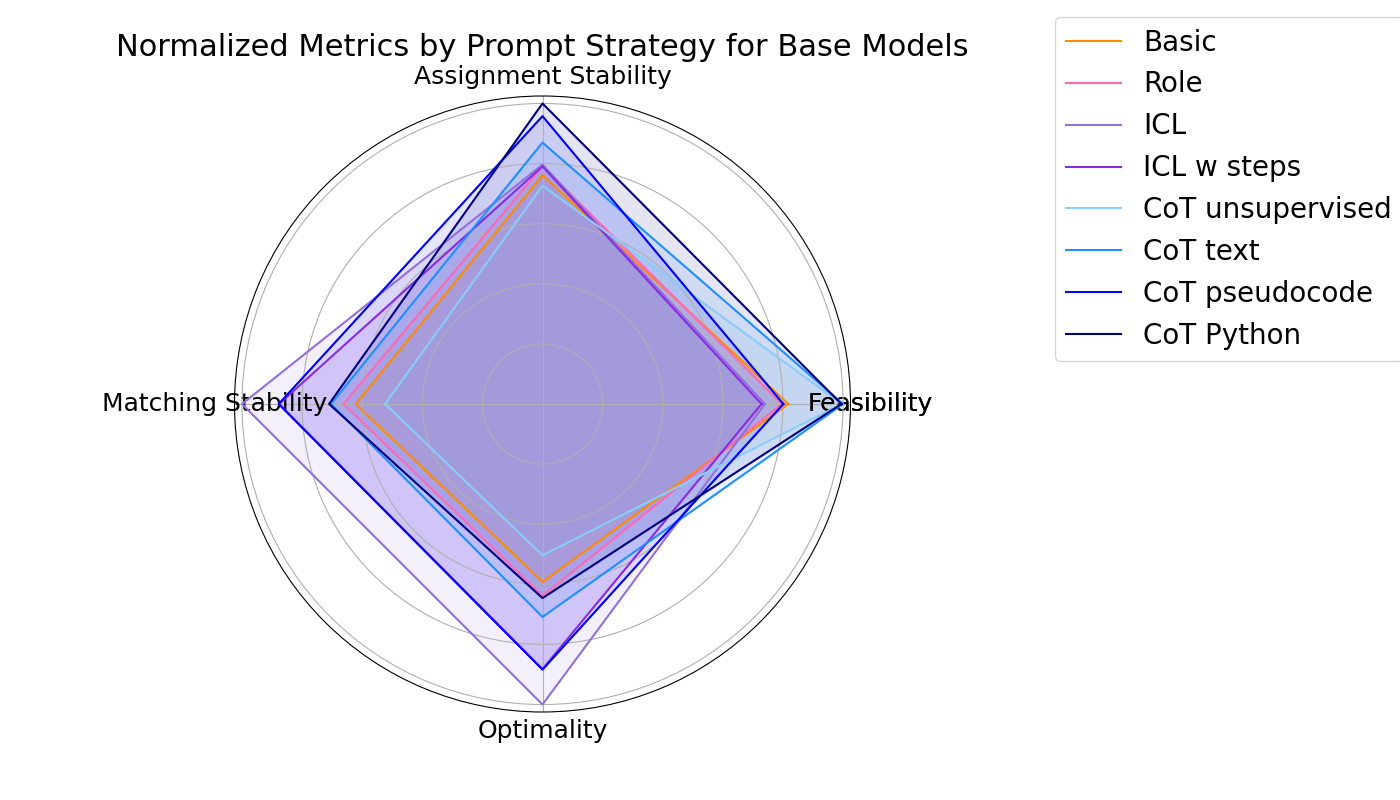}
    \caption{Performance of Prompts in Base Models}
    \label{fig:prompt_spider_base}
  \end{subfigure}
  %\hfill
  \hspace{1cm}
  \begin{subfigure}{0.45\textwidth}
    \includegraphics[width=\linewidth]{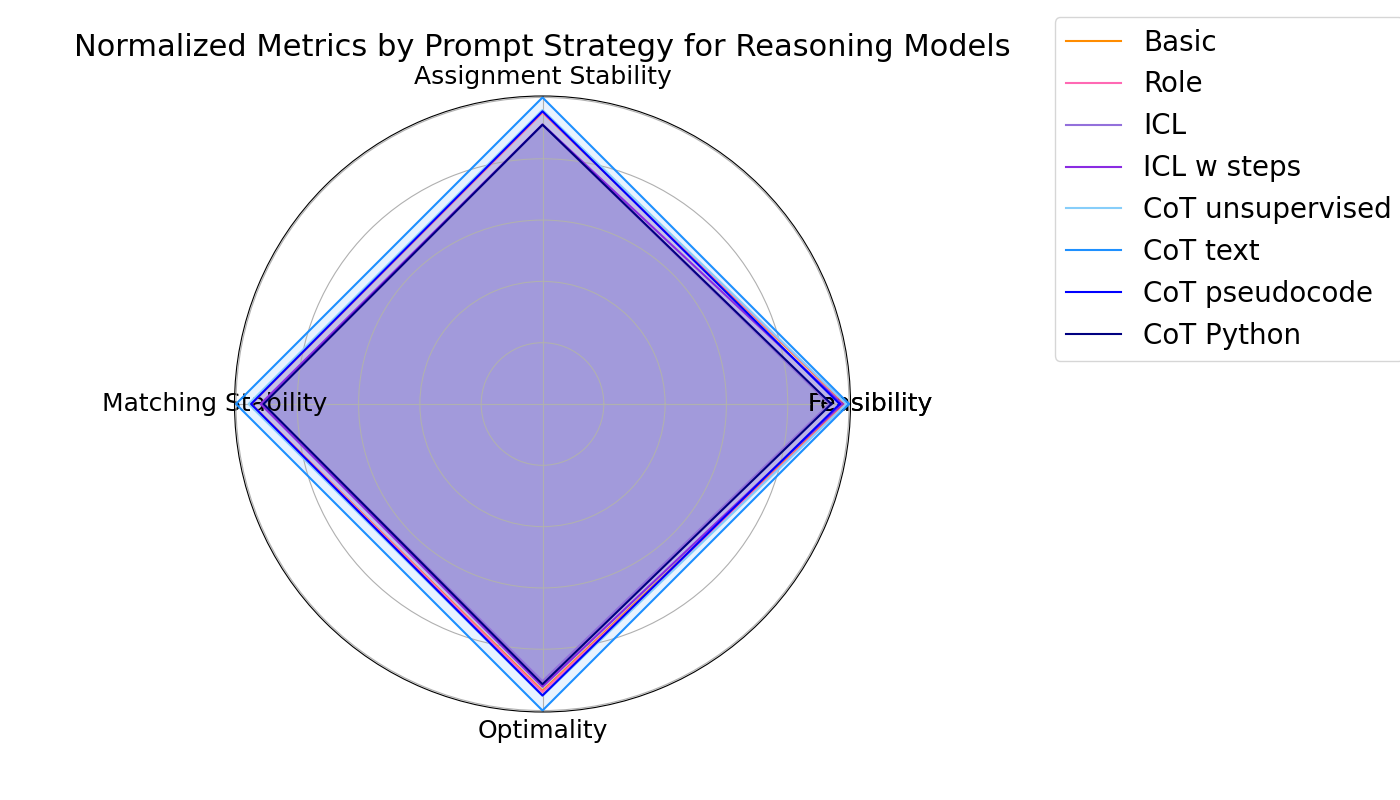}
    \caption{Performance of Prompts in Reasoning Models}
    \label{fig:prompt_spider_reas}
  \end{subfigure}

  \caption{Performance by Model Types and Performance by Prompting Styles.}
\label{fig:spider_prmpt}
\end{figure*}

In this paper, we tested multiple prompt strategies known to improve reasoning. In the Appendix~\ref{tables_graphs}, Table~\ref{overall_prompt_model} presents the metrics for each model and prompt. For the base models, the gap for a metric between two prompts can reach up to 40\%.
%where performances can be very low, the prompting strategy can have a large impact of over 40\%. 
While reasoning models seem less impacted by the choice of prompt, hinting that their reasoning is advanced enough that they do not need to rely on CoT or ICL, the difference between strategies becomes more pronounced when looking for more challenging instances. Interestingly enough, Figure~\ref{fig:spider_prmpt} shows that for the base models, the best prompts, mainly the ICL ones and CoT with pseudocode, perform worst on every metric for reasoning models. For the base LLMs, CoT unsupervised performed poorly, consistent with prior work that reported a reduction in overall performance with this strategy \citep{zhang2025prompt}. This same strategy, along with the CoT-text strategy, which performed similarly, are best suited for reasoning models. 
This suggests that we need to adapt the prompt strategy for the model used, and that LLMs benefit from having less detailed information as they develop reasoning capabilities to exploit thought processing. Ultimately, there is no single golden rule that determines which prompts are associated with higher performance for all models. 

\paragraph{RQ3: Can iterative prompting help LLM achieve higher performances through self-verification and correction?}
For the iterative prompting experiments, as we had to respect the token limits of all open-source LLMs, we restricted our experiments to 5 and 10 students. Table~\ref{iterative_prompting_by_model} presents the results under role prompting, where the same conclusions can be draw from other prompt under Appendix~\ref{tables_graphs} with Table~\ref{iterative_prompting_by_model_basic}-\ref{iterative_prompting_by_model_pseudo}. 
Additionally, while GPT-oss showed great results previously, the model is even more restrictive in the resources, so we limited the testing on the reasoning model to QwQ. 
%Table~\ref{iterative_prompting_by_model} presents the results of iterative prompting. 
%The approach was described in Section~\ref{prompt}, but we intentionally didn't mention what the final response would be in case it didn't reach the correct answer in N attempts. 
%If we return the best match, which is the one that satisfied the most metrics, the results can be significantly higher.
%, even if base LLMs do not reach the performances of reasoning LLMs without this technique. 
In the cases where the model failed to produce the correct answer within the 5 attempts, we select the best attempt, i.e., the one that satisfied the highest number of metrics.
Although this could give the impression that LLMs can understand the feedback given on this problem and correct the past errors to some degree, we can show that improvement is not monotonic by returning the last matching. In this case, while it is often possible to have very similar performance when returning the last or best attempt, there are other cases where it can also decrease significantly. This reveals that iterative prompting is not really iteratively improving the model's behavior and reasoning, but rather that generating multiple distinct answers is more beneficial in this scenario. Moreover, as the difference can only occur when the LLM fails to find the real student-optimal solution within 5 attempts, it is more challenging to evaluate the impact of self-correctness from the reasoning models. 

\begin{table}[ht]
\begin{tabular}{llrrrr}
\toprule
\small
Model & Attempt & Feasibility & Assignment & Matching & Optimality \\
\midrule
Llama 8B & No   & \prop{0.4678}  (-)  & \prop{0.117}   (-) & \prop{0.0409}  (-)  & \prop{0.0351} (-)  \\
Llama 8B & Last & \prop{0.438596} (\textcolor{red}{-2.9\%}) & \prop{0.122807} (\textcolor{DarkGreen}{+0.6\%}) & \prop{0.070175} (\textcolor{DarkGreen}{+2.9\%}) & \prop{0.05848} (+\textcolor{DarkGreen}{+2.3\%}) \\
Llama 8B & Best & \prop{0.602339} (\textcolor{DarkGreen}{+13.5\%}) & \prop{0.146199} (\textcolor{DarkGreen}{2.9\%})& \prop{0.081871} (\textcolor{DarkGreen}{+4.1\%})& \prop{0.058480} (\textcolor{DarkGreen}{+2.3\%}) \\
\midrule
Llama 70B & No   & \prop{0.6316}  (-)  & \prop{0.0819} (-)  & \prop{0.0468} (-)  & \prop{0.0234} (-) \\
Llama 70B & Last & \prop{0.853801} (\textcolor{DarkGreen}{+22.2\%}) & \prop{0.116959}  (\textcolor{DarkGreen}{+3.5\%})& \prop{0.111111} (\textcolor{DarkGreen}{+6.4\%}) & \prop{0.081871} (\textcolor{DarkGreen}{+5.6\%}) \\
Llama 70B & Best & \prop{0.888889} (\textcolor{DarkGreen}{+25.7\%}) & \prop{0.140351}  (\textcolor{DarkGreen}{+5.9\%})& \prop{0.140351} (\textcolor{DarkGreen}{+9.4\%}) & \prop{0.081871} (\textcolor{DarkGreen}{+5.6\%})\\
\midrule
Mistral 7B& No   & \prop{0.2242} (-)  & \prop{0.5614} (-)  & \prop{0.0585} (-)  & \prop{0.0585} (-) \\
Mistral 7B& Last & \prop{0.192982} (\textcolor{red}{-3.1\%}) & \prop{0.567251} (\textcolor{DarkGreen}{+0.6\%}) & \prop{0.081871}(\textcolor{DarkGreen}{+2.3\%}) & \prop{0.081871}(\textcolor{DarkGreen}{+2.3\%}) \\
Mistral 7B& Best & \prop{0.228070} (\textcolor{DarkGreen}{+0.4\%}) & \prop{0.614035} (\textcolor{DarkGreen}{+5.3\%}) & \prop{0.081871} (\textcolor{DarkGreen}{+2.3\%}) & \prop{0.081871} (\textcolor{DarkGreen}{+2.3\%}) \\
\midrule
Qwen 7B& No   & \prop{0.2339} (-)   & \prop{0.2632} (-)   & \prop{0.0351} (-) & \prop{0.0351} \\
Qwen 7B& Last & \prop{0.438596} (\textcolor{DarkGreen}{+20.5\%}) & \prop{0.076023} (\textcolor{red}{-18.7\%})& \prop{0.052632} (\textcolor{DarkGreen}{+1.8\%}) & \prop{0.046784} (\textcolor{DarkGreen}{+1.2\%}) \\
Qwen 7B& Best & \prop{0.549708} (\textcolor{DarkGreen}{+31.6\%}) & \prop{0.19883} (\textcolor{DarkGreen}{-6.4\%}) & \prop{0.05848} (\textcolor{DarkGreen}{+2.3\%})  & \prop{0.046784} (\textcolor{DarkGreen}{+1.2\%}) \\
\midrule
QwQ 32B& Non   & \prop{1}   (-)    & \prop{0.9181} (-)   & \prop{0.9181} (-)  & \prop{0.8304} (-)   \\
QwQ 32B& Last & \prop{1.0}   (-)  & \prop{0.994152} (\textcolor{DarkGreen}{+7.6\%}) & \prop{0.994152} (\textcolor{DarkGreen}{+7.6\%}) & \prop{0.953216} (\textcolor{DarkGreen}{+12.3\%})\\
QwQ 32B& Best & \prop{1.0}  (-)   & \prop{0.988304} (\textcolor{DarkGreen}{+7.0\%}) & \prop{0.988304} (\textcolor{DarkGreen}{+7.0\%})& \prop{0.94152} (\textcolor{DarkGreen}{+11.1\%}) \\

\bottomrule
\end{tabular}
\caption{Iterative prompting with Role prompting for instance with 5-10 students. For each model, we have the metrics for no iterative prompting and iterative prompting with the last and best attempt}
\label{iterative_prompting_by_model}
\end{table}

\section{Discussion, Limitations and Future Work} \label{discu}

The results of RQ1 showed that the many-to-one matching problem is a challenging task, where LLMs fail to satisfy the notions of stability, feasibility, and optimality, regardless of the prompts, as the number of students increases. Indeed, they indicate that certain aspects of the DA algorithm's runtime complexity, which expand the solution space, such as an increase in the number of students, are associated with a sharp decline in performance. However, the impact is not the same as the preferences, which also impact the algorithm's complexity and solution space. LLMs do not have a problem with incomplete preferences, which create invalid pairs, but it also does not lead to significantly better performance. One possible hypothesis is that when students are matched early in the iterative process, the distinction between preference types has limited impact, whereas the number of students consistently influences model performance.
However, complete preferences may introduce greater difficulty in instances where the student-optimal matching assigns students to colleges ranked among their least preferred options. More experiments are needed to analyze how preference constraints impact the solution when scaling, or whether they would still lead to the same conclusions if the models were not instructed to follow this algorithm, where invalid pairs are naturally avoided. 

Moreover, the results of RQ2 highlight the sensitivity of the models. Indeed, we observe that there is no prompt which systematically offers better performances over all models. More precisely, the performances under a prompting strategy depend on several aspects, including the model used, the corresponding number of parameters it has, and the observed metric. 
%Our results suggest that a more in-depth investigation of these instabilities is needed to better understand which promoting strategies are the most appropriate to various problem conditions. 
Additionally, even if we attempt to enhance the performance of base LLMs with these strategies, they will still fall short of matching the performance of reasoning LLMs with minimal prompting. However, there is a trend between strategies that are better fitted for base or reasoning LLMs. We hypothesize that base models can benefit from having detailed prompts, which will guide their thought process, making a substantial difference in the final results, while reasoning models have sufficient capabilities to do so with general guidelines.

Additionally, RQ3 demonstrates that iterative prompting is a promising approach for improving LLMs' reasoning under challenging tasks. However, the fact that results do not converge iteratively toward a better solution raises questions regarding the actual capacity of LLMs to integrate feedback and use it to improve their answers. Indeed, more experiments are necessary to determine whether the performances increase as a result of potential self-correction or simply by an increase in the answer diversity. An interesting direction could be to experiment with different feedback forms that vary in the level of detail provided regarding the encountered errors, like specifying the exact unstable pairs rather than only reporting the total number of unstable pairs in the matching.

While this work provides a valuable benchmark for a previously underexplored setting, it still has some limitations. Although testing each strategy allowed us to observe its individual impact on performance, further improvements could be achieved by combining strategies or by extending our current designs to include multiple examples in ICL. A common challenge in evaluating the most accessible LLMs, such as Llama, is fitting the prompt and output within the limit of allowed tokens, which also prevent us from scaling the instances to bigger sizes. Another limitation concerns the models tested in this article. Although other LLMs have achieved stronger performance on reasoning tasks, we limited our study to open-source models that can be used with reasonable resources. 
While future work could address current limitations, this work could also serve as an initial step toward aligning optimization outcomes with individual preferences, a concept often referred to as preferential alignment. While the DA algorithm guarantees stability through a strategy-proof mechanism, it can be inefficient \citep{ortega2023cost}. Alternatives like Top Trading Cycles (TTC) and Risk Minimization (RM) offer solutions that may lead to higher overall satisfaction, even though they may produce matchings that are not stable \citep{ortega2023cost}. Therefore, depending on the application, some users may prefer an outcome that prioritizes achieving their first rank through a highly efficient approach. In contrast, others may prefer stability, strategy-proofness, or fairness. LLMs offer a promising complement: not only as solvers but as interpreters capable of suggesting appropriate approaches based on the context and users' priorities through even deeper reasoning.

\section{Conclusions} \label{conclusion}
Solving a two-sided many-to-one matching problem remains highly challenging for LLMs. Although an algorithm exists that can compute a feasible, stable, and student-optimal solution in polynomial time, ensuring that the outputs of the models respect these properties is far from straightforward. Our benchmark reveals how well models handle structured preference-based reasoning, providing direct insight into their ability to align with domain-specific constraints. Specific properties, such as achieving a student-optimal stable matching, are particularly difficult to attain. Our iterative prompting experiments highlight the core challenge of self-correction based on feedback. We also observed that the models are more significantly affected by the number of students than by the type of preferences, even though both factors impact the running time of the DA algorithm. This suggests that the models struggle to follow the iterative structure required by this polynomial-time procedure correctly. Reasoning LLMs consistently outperform base ones across all metrics. Surprisingly, Mistral shows strong performance on stability despite being a base model. While no prompting strategy is universally optimal across all models, each model has a CoT variant that performs reliably well, making it a broadly effective approach.

\section*{Acknowledgements}
Funding support for project activities has been partially provided by the Canada CIFAR AI Chair, IVADO, and CIFAR Catalyst award. We also thank Compute Canada and Mila clusters for their support in providing facilities for our evaluations. This work was funded by the NSERC Grant No. 2024-04051, and the Canada Graduate Research Scholarship. It is also funded by the FRQNT Master Scholarship No. B1X-341759. 

\bibliography{ref}
\bibliographystyle{plain}

%%%%%%%%%%%%%%%%%%%%%%%%%%%%%%%%%%%%%%%%%%%%%%%%%%%%%%%%%%%%

\appendix

\section{Instances and Prompts} \label{instance_prompt}
An instance of the College Admission Problem consists of students and colleges that must be written in a specific format to be compatible with the DA algorithm. Therefore, we introduce in the following Figure~\ref{prompt:instance_ex} an example of such an instance. Specifically, each instance consists of the number of students and colleges, followed by a list stating their names. The capacities of each school will follow it, and then we will have the preferences of students and colleges, respectively. As we compare the performance of the output of LLMs with the results of the DA algorithm, we compile the matching given by the DA algorithm for this prompt. The Figure~\ref{prompt:DA_ex} provides the answer format, as we expect it to be returned by LLM models. Note that when a student is matched to "nothing", it simply means that this student is unassigned. There is a pair for each student in the matching, but it is not necessarily for each school if no students are assigned to that school. 

\begin{figure}
\begin{promptbox} 
\# Num. students:10 \\
\# Num. colleges:3 \\
\# Students:s1,s2,s3,s4,s5,s6,s7,s8,s9,s10 \\
\# Colleges:c1,c2,c3 \\
\# Capacities: \\
c1 3 \\
c2 2 \\
c3 3 \\
\# Student preferences: \\
s1 (1,c3) (2,c2) (3,c1) \\
s2 (1,c1) (2,c2) (3,c3) \\ 
s3 (1,c3) (2,c1) (3,c2) \\
s4 (1,c1) (2,c3) (3,c2) \\ 
s5 (1,c3) (2,c1) (3,c2) \\
s6 (1,c3) (2,c1) (3,c2) \\
s7 (1,c3) (2,c1) (3,c2) \\
s8 (1,c3) (2,c2) (3,c1) \\
s9 (1,c3) (2,c2) (3,c1) \\
s10 (1,c2) (2,c3) (3,c1) \\
\# College priorities: \\
c1 (1,s2) (2,s1) (3,s10) (4,s9) (5,s7) (6,s6) (7,s4) (8,s3) (9,s5) (10,s8) \\
c2 (1,s9) (2,s8) (3,s6) (4,s5) (5,s10) (6,s1) (7,s4) (8,s3) (9,s2) (10,s7) \\
c3 (1,s8) (2,s1) (3,s6) (4,s3) (5,s7) (6,s10) (7,s5) (8,s9) (9,s4) (10,s2) \\
\end{promptbox}
\caption{Prompt example of an College Admission instance.}
\label{prompt:instance_ex}
\end{figure}

\begin{figure}
\begin{promptbox}
\# DA matching of corresponding input  

[("s1", "c3"), ("s2", "c1"), ("s3", "nothing"), ("s4", "nothing"), ("s5", "c2"), ("s6", "c3"), ("s7", "c1"), ("s8", "c3"), ("s9", "c2"), ("s10", "c1")]
\end{promptbox}
\caption{Example of the results proposed by the DA algorithm.}
\label{prompt:DA_ex}
\end{figure}

As detailed in Section~\ref{benchmark_details}, our benchmark evaluates model performance across four levels of student population, three types of preference, 3 different capacity level, four ratios and three random seeds. Certain combinations are excluded due to infeasibility, like under-capacity settings with a 1:1 ratio, which would imply fewer seats than schools, which is not meaningful in this context. The final dataset comprises 369 total instances per model–prompt combination.  Specifically, for each preference type, there are 123 instances. For each model-prompt combination, we have 99 instances corresponding to 10,15 or 20 students, while there are 72 for 5 students. This design enables fine-grained analysis across different parameter settings while ensuring a sufficient number of instances to support statistically meaningful conclusions.

Each of the 369 benchmark instances was evaluated across six language models and eight distinct prompting strategies, as described in Section~\ref{prompt}. Following conventions established in prior work  \citep{juneja2025task, liu2025beyond}, each prompt was structured into distinct components, with explicit headers indicating the purpose of each section. We also appended the phrase "Final Matching:" at the end of each prompt to explicitly signal the model to return the final assignment. This technique, which is commonly used in question-answering tasks, helps improve output validity \citep{zhang2023automatic}. The prompts are presented from Figure~\ref{basic_prompt} to Figure~\ref{python_CoT_prompt}, with redundant content omitted for brevity.

\begin{figure*}[] 
\begin{promptbox_whole} 
\small
Task

Implement the student-proposing deferred acceptance algorithm (Gale-Shapley) to solve a many-to-one matching problem for college admissions.
\\

Instance Format

The instance will contain the following elements: 
\begin{itemize}
    \item Num. students: Integer, the number of students
    \item Num. colleges: Integer, the number of colleges
    \item Students: List of student names
    \item Colleges: List of college names
    \item Capacities: Dictionary, each college mapped to its capacity (integer)
    \item Student preferences: Dictionary, each student is mapped to a list of tuples (rank,college) 
    \item College preferences: Dictionary, each college is mapped to a list of tuples (rank, student)
\end{itemize}
Ranks are integers where 1 indicates the highest preference.
\\

Constraints

The output matching must satisfy the following:

Feasibility: 
\begin{itemize}
    \item Each student is matched to at most one college.
    \item No college is matched with more students than its capacity.
\end{itemize}

Stability: 

\begin{itemize}
    \item [] There must be no blocking pair (student, college) such that: 
    \end{itemize}
    \begin{itemize}
        \item The student prefers the college over their current match, and
        \item The college either has not reached its capacity or prefers the student to at least one of its current matches.
        \end{itemize}
        
Optimality: 
\begin{itemize}
    \item []  Among all stable matchings, students are matched to the best possible college based on their preferences.
    \end{itemize}
\medskip

Output Format

Return the matching as a list of tuples of the form (student, match). If a student is unmatched, use “nothing” as the college. 
\\

Instance

The instance to solve is:
[ADD INSTANCE HERE]
\\

Instruction

Return only the final matching as a list of tuples.
\\

Final Matching: 

\end{promptbox_whole}
\caption{Template of our \textbf{Basic} prompt strategy}
\label{basic_prompt}
\end{figure*}

\begin{figure*}[] 
\begin{promptbox_whole} 
\small
You are an optimization algorithm expert specializing in stable matching algorithms, particularly the Gale-Shapley deferred acceptance algorithm.
\\

Task\\
...
\\

Instance Format\\
...
\\

Constraints\\
...
\\

Output Format\\
...
\\

Instance\\
...
\\

Instruction\\
...
\\

Final Matching: 

\end{promptbox_whole}
\caption{Template of our \textbf{Role} prompt strategy}
\label{role_prompt}
\end{figure*}

\begin{figure*}[] 
\begin{promptbox_whole} 
\small
Task \\
...
\\

Instance Format\\
...
\\

Constraints \\
...
\\

Output Format\\
...
\\

Example

Below is a solved example demonstrating input format and expected output. 
[ADD EXAMPLE HERE]
\\

Instance \\
...
\\

Instruction\\
...
\\

Final Matching: 

\end{promptbox_whole}
\caption{Template of our \textbf{ICL} prompt strategy}
\label{icl_prompt}
\end{figure*}

\begin{figure*}[] 
\begin{promptbox_whole} 
\small
Task

...

Instance Format\\
...

Constraints\\
...

Output Format\\
...

Example

Here is an example consisting of an instance, the step-by-step application of the algorithm, and the final matching.

**Instance** \\
\# Num. students: 5 \\
\# Num. Colleges: 3 \\
\# Students: s1, s2, s3, s4, s5 \\
\# Colleges: c1, c2, c3\\
\# Capacities: \\
c1  2,\\
c2  1,\\
c3  2 \\
\# Student preferences: \\
s1 (1,c1) (2,c2) (3,c3)\\
s2 (1,c1) (2,c3) (3,c2)\\
s3 (1,c2) (2,c1) (3,c3)\\
s4 (1,c3) (2,c2) (3,c1)\\
s5 (1,c1) (2,c2) (3,c3)\\
\# Colleges priorities: \\
c1 (1,s2) (2,s1) (3,s5) (4,s3) (5,s4)\\
c2 (1,s3) (2,s1) (3,s4) (4,s5) (5,s2)\\
c3 (1,s4) (2,s2) (3,s5) (4,s1) (5,s3)\\
**Step by Step DA Algorithm**\\
Round 1 : 
\begin{itemize}
    \item[] s1, s2, s3, s4, s5 propose to their 1st choices, c1, c1, c2, c3, c1 respectively. 
    \item[] c1 (capacity of 2) gets proposals from s1, s2, s5. It prefers s1 and s2. It rejects s5.
    \item[] c2 (capacity of 1) gets a proposal from s3 and holds it
    \item[] c3 (capacity of 2) gets a proposal from s4 and holds it.
    \item[] Applications of hold :  c1 :{s1,s2}, c2 :{s3} and c3:{s4}. Rejected:{s5}
    \end{itemize}
Round 2 : 
\begin{itemize}
    \item[] s5 (previously rejected) proposes to its next choice, c2.
    \item[] c2 (capacity of 1) compares the new proposal (s5) to its current hold (s3). It prefers s3. It rejects s5.
    \item[] Applications of hold :  c1 :{s1,s2}, c2 :{s3} and c3:{s4}. Rejected:{s5}
    \end{itemize}
Round 3 : 
\begin{itemize}[ label={}, itemsep=0pt, topsep=0pt, parsep=0pt, partopsep=0pt]
    \item s5 (previously rejected) proposes to its next choice, c2.
    \item c3 (capacity of 2) gets a proposal from s5. It is holding s4 and has an open spot. It holds s5.
    \item Applications of hold :  c1 :{s1,s2}, c2 :{s3} and c3:{s4,s5}. Rejected:{}
    \end{itemize}
There was no rejection on the last round, the algorithm ends and the colleges accept the students on hold. \\
**Final Matching** \\
The final matching is : \\
(s1,c1),(s2,c1),(s3,c2),(s4,c3),(s5,c3)\\

Instance\\
...

Instruction \\
...

Final Matching: 

\end{promptbox_whole}
\caption{Template of our \textbf{ICL w steps} prompt strategy}
\label{ICL_steps_prompt}
\end{figure*}

\begin{figure*}[] 
\begin{promptbox_whole} 
\small
Task\\
...
\\

Instance Format\\
...
\\

Constraints\\
...
\\

Instance \\
...
\\

Instruction \\
Return only the final matching as a list of tuples. Think step by step. 
\\

Final Matching: 
\end{promptbox_whole}
\caption{Template of our \textbf{CoT unsupervised} prompt strategy}
\label{unsup_CoT_prompt}
\end{figure*}

\begin{figure*}[] 
\begin{promptbox_whole} 
\small
Task\\
...
\\

Instance Format\\
...
\\

Constraints\\
...
\\

Output Format\\
...
\\

Instance \\
...
\\

Step by Step Process\\
\begin{enumerate}
    \item Each student applies to their favorite college. 

    \item Each college rejects all applications from students that are unacceptable to it. If a college received at most q applications from acceptable students so far, all those students are put on the colleges waiting list. Otherwise the college puts its favorite q students among all applicants on the waiting list and rejects all remaining ones. 

    \item Each student that was rejected in the previous step applies to his favorite among the colleges he or she has not yet applied to. 

    \item Steps 2 and 3 are repeated until it holds for all students that they were either not rejected in the previous step or already applied to all colleges acceptable to them.

    \item Each college admits all students on its waiting list.
    \end{enumerate}

Instruction\\
...
\\

Final Matching: 
\end{promptbox_whole}
\caption{Template of our \textbf{text CoT} prompt strategy}
\label{txt_CoT_prompt}
\end{figure*}

\begin{figure*}[] 
\begin{promptbox_whole} 
\small
Task\\
...
\\

Instance Format\\
...
\\

Constraints\\
...
\\

Output Format\\
...
\\

Instance \\
...
\\
Step by Step Process
\begin{lstlisting}[
  basicstyle=\ttfamily\small,
  frame=single,
  commentstyle=\color{gray},
  keywordstyle=\color{blue},
  literate={←}{{$\gets$}}1 {∪}{{$\cup$}}1 {∅}{{$\emptyset$}}1 {≤}{{$\leq$}}1
]
# INITIALIZE:
for each student s in S:
  s.next_choice ← 1        # index into Pref(s), starting at top choice
  s.is_free     ← true
for each college c in C:
  c.waiting_list ← ∅       # no one tentatively held yet
  c.applicants   ← ∅       # applications in the current round

REPEAT
  # Free students apply to their next choice
  for each student s in S:
    if s.is_free and s.next_choice ≤ length(Pref(s)): 
      let c = Pref(s)[s.next_choice]
      add s to c.applicants
      s.next_choice ← s.next_choice + 1

  # Colleges review their applicant pools
  any_rejection ← false
  for each college c in C:
    pool ← c.waiting_list ∪ c.applicants
    remove from pool any student unacceptable to c
    sort pool by c's priority order (best first)
    c.waiting_list ← first q(c) students in pool
    rejected ← pool minus c.waiting_list
    for each student r in rejected:
      r.is_free ← true
      any_rejection ← true
    for each student a in c.waiting_list:
      a.is_free ← false
    c.applicants ← ∅

UNTIL any_rejection is false

# Final assignments
for each college c in C:
  admit every student in c.waiting_list
\end{lstlisting}

Instruction \\ 
...
\\

Final Matching:

\end{promptbox_whole}
\caption{Template of our \textbf{pseudocode CoT} prompt strategy}
\label{pseudo_CoT_prompt}
\end{figure*}

\begin{figure*}[] 
\begin{promptbox_whole} 
\small
Task\\
...
\\

Instance Format\\
...
\\

Constraints\\
...
\\

Output Format\\
...
\\

Instance \\
...
\\

Step by Step Process
\begin{lstlisting}[
    language=Python,
    basicstyle=\ttfamily\small,
    frame=single,
    keywordstyle=\color{blue},
    commentstyle=\color{gray},
    stringstyle=\color{orange},
    showstringspaces=false,
    tabsize=4,
    breaklines=true
]
priority_ranks = {
    c: {s: rank for rank, s in enumerate(priority)}
    for c, priority in college_priorities.items()
}

proposals = {s: deque(prefs) for s, prefs in student_prefs.items()}
college_matches = defaultdict(list)
student_match = {s: None for s in students}

while True:
    free_students = [
        s for s in students 
        if student_match[s] is None and proposals[s]
    ]
    if not free_students:
        break

    for student in free_students:
        college = proposals[student].popleft()
        college_matches[college].append(student)

        accepted = sorted(
            college_matches[college],
            key=lambda s: priority_ranks[college][s]
        )[:capacities[college]]

        for s in college_matches[college]:
            if s in accepted:
                student_match[s] = college
            else:
                student_match[s] = None

        college_matches[college] = accepted
\end{lstlisting}

Instruction \\ 
...
\\

Final Matching:

\end{promptbox_whole}
\caption{Template of our \textbf{Python CoT} prompt strategy}
\label{python_CoT_prompt}
\end{figure*}

\paragraph{Iterative Prompting Feedback:}
For iterative prompting, we are providing general feedback on the solution returned by the LLM. More specifically, if there is a matching in the previous output, we will return the feedback following this example that was feasible and stable, but not optimal.
\begin{figure}
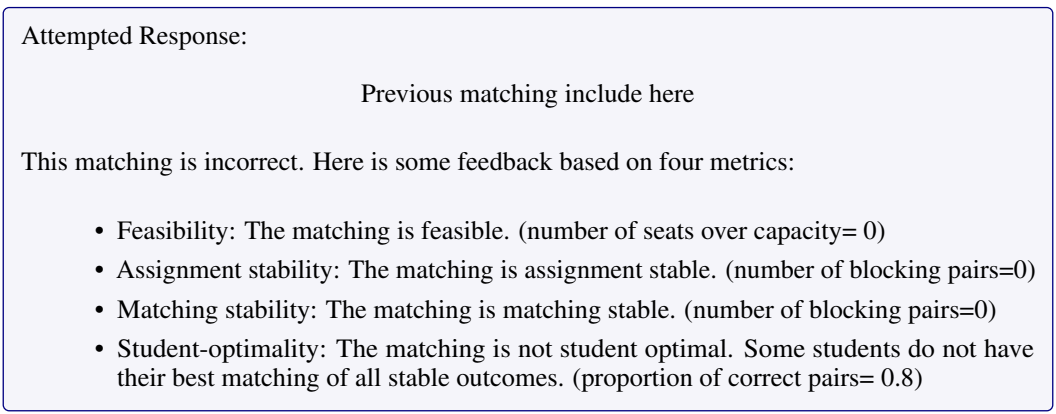

\begin{promptbox}
    Attempted Response: \\
    \[ \text{Previous matching include here} \] \\
    This matching is incorrect. Here is some feedback based on four metrics: \\
    \begin{itemize}
        \item Feasibility: The matching is feasible. (number of seats over capacity= 0)
        \item Assignment stability: The matching is assignment stable. (number of blocking pairs=0)
        \item Matching stability: The matching is matching stable. (number of blocking pairs=0)
        \item Student-optimality: The matching is not student optimal. Some students do not have their best matching of all stable outcomes. (proportion of correct pairs= 0.8)
    \end{itemize}
    
\end{promptbox}
\caption{Feedback template example for a feasible and stable, but not student-optimal matching with iterative prompting.}
\label{prompt:iterative_feedback}
\end{figure}

\section{Valid Output} \label{valid}
We defined validity in Section~\ref{evaluation}. Table~\ref{count_table} reports the proportion of the 369 instances that are valid. Invalid outputs include cases where the generated matching contains incorrect names (e.g., referring to students as "Alice" and "Bob" instead of s1 and s2), or when the model returns code rather than a matching. 
As shown in Table~\ref{count_table}, there is no major problem and the proportions of valid output are almost always close to 1. However, Mistral exhibits a comparatively higher rate of invalid outputs, particularly when using the ICL with steps prompt. Qwen also produces a smaller number of invalid outputs under certain CoT prompting strategies. In contrast, both Llama models consistently generate valid outputs across all instances. 
It is worth noting that most invalid responses occurred for larger instances, suggesting that model can have a harder time following the instruction with bigger instances. 

\begin{table}[]
\begin{tabular}{l|llllll}
      & Llama 8B & Llama 70B & Mistral  & Qwen & QwQ & GPT-oss \\ \hline
Basic &   \prop{1}    &   \prop{1}     & \prop{0.9837398}  &\prop{1}  &\prop{0.989159892} & \prop{0.991869919}\\
Role  &    \prop{1}   &  \prop{1}      & \prop{0.98644986} &\prop{1}  &\prop{0.99728997} & \prop{0.991869919}  \\
ICL   &    \prop{1}   &   \prop{1}     & \prop{0.785907859} & \prop{1}  &\prop{0.97831978} & \prop{0.994579946} \\
ICL w steps  & \prop{0.945799458} &   \prop{1}    & \prop{0.97289973} &  \prop{1}   & \prop{0.97289973}&  \prop{0.99728997}  \\
CoT unsupervised &  \prop{1}   &  \prop{1} & \prop{0.956639566}   & \prop{0.99728997}& \prop{1} & \prop{0.994579946} \\
CoT text & \prop{1}  &   \prop{1}    & \prop{0.9837398}    &  \prop{0.880758808} & \prop{0.994579946}& \prop{0.989159892}\\
CoT pseudocode &  \prop{1}  &  \prop{1}   & \prop{0.970189702} & \prop{0.986449864} & \prop{0.994579946}& \prop{0.99728997} \\
CoT Python & \prop{1}  & \prop{1}  & \prop{0.967479675}  & \prop{1} &\prop{0.9837398}  & \prop{0.99728997}\\
\end{tabular}
\caption{Percentage of output, out of the 369 possibilities, that generate an acceptable matching (i.e a matching that we could extract from the output)}
\label{count_table}
\end{table}

\section{Tables and Graphs}\label{tables_graphs}
\subsection{Time Table}
We present the time each LLM took in order to generate the output of 10 random instances across our datasets, making sure we had similar numbers for different complexity. 
\begin{table}[ht]
\begin{tabular}{llrrrrr}
\toprule
\small
          Model &    Llama 8B &  Llama 70B &  Mistral 7B &  Qwen 7B &  QwQ 32B  &  GPT-oss 120B \\
\midrule
                      Average Time &  14.3111 &6.9977 & 1.7036 &  6.5190 & 391.7150 & 324.5929 \\
                      Standard deviation &5.347 & 2.918 & 0.9432 &4.930 &129.0224 & 308.9703 \\
\bottomrule
\end{tabular}
\caption{Average time (in seconds) to generate the output for the College Admission Problem across 10 intances. }
\label{time_table}
\end{table}

\subsection{Aggregated Tables}
We include in this section, under Table~\ref{overall_prompt_model} the aggregated results across the 369 instances, when are of them are valid, for each prompt strategy and model combination. 
While our figures included in the paper represent very well the data by number of students and preference types, this table allows to have a more broad view of the overall performances. 
\begin{table}[ht]
\begin{tabular}{llrrrrr}
\toprule
\small
          Prompt &    Model &  Count &  Feasibility &  Assignment &  Matching  &  Optimality \\
\midrule
                      Basic &  Llama8B & 369 & \prop{0.349593} & \prop{0.054201} & \prop{0.016260} & \prop{0.013550}   \\
                      Basic & Llama70B & 369 & \prop{0.355014} & \prop{0.146341} & \prop{0.024390} & \prop{0.013550}     \\
            Basic &  Mistral & 363 & \prop{0.146006} & \prop{0.479339} & \prop{0.030303} & \prop{0.030303} \\
            Basic &     Qwen & 369 & \prop{0.170732} & \prop{0.108401} & \prop{0.010840} & \prop{0.010840} \\
            
            Basic &     QwQ & 365 & \prop{0.875} & \prop{0.646739} & \prop{0.633152} & \prop{0.578804} \\
            
  Basic &     GPT-oss & 366 & \prop{0.983607} & \prop{0.811475} & \prop{0.8060111} & \prop{0.792350} \\
           \midrule
                        Role &  Llama8B & 369 & \prop{0.352304} & \prop{0.054201} & \prop{0.018970} & \prop{0.016260}  \\
             Role & Llama70B & 369 & \prop{0.336043} & \prop{0.168022} & \prop{0.021680} & \prop{0.010840} \\
            Role &  Mistral & 364 & \prop{0.159341} & \prop{0.456044} & \prop{0.030220} & \prop{0.030220} \\
            Role &     Qwen & 369 & \prop{0.146341} & \prop{0.140921} & \prop{0.016260} & \prop{0.016260} \\
            Role &     QwQ & 368 & \prop{0.875}    & \prop{0.646739} & \prop{0.633152} & \prop{0.578804} \\
  Role &     GPT-oss & 366 & \prop{0.989071} & \prop{0.811475} & \prop{0.806011} & \prop{0.797814} \\
            \midrule
            ICL &  Llama8B & 369 & \prop{0.208672} & \prop{0.086721} & \prop{0.027100} & \prop{0.021680}      \\
 ICL & Llama70B & 369 & \prop{0.330623} & \prop{0.208672} & \prop{0.040650} & \prop{0.029810} \\
            ICL &  Mistral & 290 & \prop{0.189655} & \prop{0.458621} & \prop{0.044828} & \prop{0.044828} \\
  ICL &     Qwen & 369 & \prop{0.195122} & \prop{0.070461} & \prop{0.018970} & \prop{0.018970} \\
            ICL &     QwQ & 361 & \prop{0.775623} & \prop{0.590028} & \prop{0.578947} & \prop{0.531856} \\
                        ICL &     GPT & 367 & \prop{0.986376} & \prop{0.80654} & \prop{0.80109} & \prop{ 0.792916} \\
             \midrule
    ICL w steps &  Llama8B & 349 & \prop{0.335244} & \prop{0.020057} & \prop{0.005731} & \prop{0.005731}   \\
    ICL w steps & Llama70B & 369 & \prop{0.279133} & \prop{0.249322} & \prop{0.040650} & \prop{0.029810} \\
     ICL w steps &  Mistral & 359 & \prop{0.108635} & \prop{0.467967} & \prop{0.044568} & \prop{0.041783} \\
     ICL w steps &     Qwen & 369 & \prop{0.186992} & \prop{0.081301} & \prop{0.024390} & \prop{0.024390} \\
     ICL w steps &     QwQ & 359 & \prop{0.852368} & \prop{0.598886} & \prop{0.598886} & \prop{0.559889} \\
     ICL w steps &     GPT & 368 & \prop{0.98913} & \prop{0.798913} & \prop{0.793478} & \prop{0.790761} \\
     \midrule
CoT unsup. &  Llama8B & 369 & \prop{0.403794} & \prop{0.046070} & \prop{0.013550} & \prop{0.010840}    \\
CoT unsup. & Llama70B & 369 & \prop{0.360434} & \prop{0.165312} & \prop{0.018970} & \prop{0.010840}   \\
CoT unsup. &  Mistral & 353 & \prop{0.172805} & \prop{0.467422} & \prop{0.025496} & \prop{0.025496} \\
CoT unsup. &     Qwen & 368 & \prop{0.307065} & \prop{0.073370} & \prop{0.010870} & \prop{0.010870} \\
CoT unsup. &     QwQ & 369 & \prop{0.894309} & \prop{0.674797} & \prop{0.661247} & \prop{0.607046} \\
CoT unsup. &     GPT & 367 & \prop{0.972752} & \prop{0.79564} & \prop{0.790191} & \prop{0.787466} \\
\midrule
        CoT text &  Llama8B & 369 & \prop{0.598916} & \prop{0.024390} & \prop{0.005420} & \prop{0.005420}  \\
        CoT text & Llama70B & 369 & \prop{0.319783} & \prop{0.184282} & \prop{0.032520} & \prop{0.021680} \\
       CoT text &  Mistral & 363 & \prop{0.151515} & \prop{0.528926} & \prop{0.033058} & \prop{0.033058} \\
        CoT text &     Qwen & 325 & \prop{0.178462} & \prop{0.163077} & \prop{0.021538} & \prop{0.021538} \\
        CoT text &     QwQ & 367 & \prop{0.896458} & \prop{0.741144} & \prop{0.724796} & \prop{0.686649} \\
        CoT text &     GPT & 365 & \prop{0.986301} & \prop{0.791781} & \prop{0.789041} & \prop{0.778082} \\
        \midrule
CoT pseudo &  Llama8B & 369 & \prop{0.382114} & \prop{0.037940} & \prop{0.008130} & \prop{0.008130}  \\
 CoT pseudo & Llama70B & 369 & \prop{0.325203} & \prop{0.146341} & \prop{0.037940} & \prop{0.027100} \\
CoT pseudo &  Mistral & 358 & \prop{0.125698} & \prop{0.650838} & \prop{0.044693} & \prop{0.044693} \\
 CoT pseudo &     Qwen & 364 & \prop{0.167582} & \prop{0.156593} & \prop{0.024725} & \prop{0.021978} \\
       CoT pseudo &     QwQ & 367 & \prop{0.841962} & \prop{0.651226} & \prop{0.634877} & \prop{0.599455} \\
              CoT pseudo &     GPT-oss & 368 & \prop{0.983696} & \prop{0.8125} & \prop{0.804348} & \prop{0.793478} \\
 \midrule
      CoT python &  Llama8B & 369 & \prop{0.601626} & \prop{0.054201} & \prop{0.013550} & \prop{0.013550}   \\
      CoT python & Llama70B & 369 & \prop{0.319783} & \prop{0.216802} & \prop{0.027100} & \prop{0.018970} \\
      CoT python &  Mistral & 357 & \prop{0.103641} & \prop{0.672269} & \prop{0.039216} & \prop{0.039216} \\
      CoT python &     Qwen & 369 & \prop{0.216802} & \prop{0.092141} & \prop{0.013550} & \prop{0.002710} \\
      CoT python &     QwQ & 363 & \prop{0.785124} & \prop{0.608815} & \prop{0.595041} & \prop{0.559229} \\
            CoT python &     GPT-oss & 368 & \prop{0.98913} & \prop{0.788043} & \prop{0.782609} & \prop{0.779891} \\
\bottomrule
\end{tabular}
\caption{Overall performances over our 4 metrics for all model and prompt. For brevity, Assignment means Assignment Stability, Matching means Matching Stability, CoT unsup. refers to CoT unsupervised and CoT pseudo refers to CoT pseudocode.}
\label{overall_prompt_model}
\end{table}

\subsection{Iterative Prompting Tables}
While Table~\ref{iterative_prompting_by_model} showed the main conclusions we can draw from iterative prompting, we present here the exact performances for role-based prompting, but also for every other prompt to show that the conclusions drawn before stay relevant with other prompts.

\begin{table}[ht]
\begin{tabular}{llrrrr}
\toprule
\small
Model & Iterative prompting & Feasibility & Assignment & Matching & Optimality \\
\midrule
Llama8B & No   & \prop{0.4561}   & \prop{0.1111}   & \prop{0.0351}   & \prop{0.0292} \\
Llama8B & Last & \prop{0.479532} & \prop{0.093567} & \prop{0.05848}  & \prop{0.046784} \\
Llama8B & Best & \prop{0.614035} & \prop{0.122807} & \prop{0.070175} & \prop{0.046784} \\

\midrule
Llama70B & No   & \prop{0.6491}   & \prop{0.0994}   & \prop{0.0526}   & \prop{0.0292} \\
Llama70B & Last & \prop{0.888889} & \prop{0.128655} & \prop{0.116959} & \prop{0.081871} \\
Llama70B & Best & \prop{0.894737} & \prop{0.152047} & \prop{0.134503} & \prop{0.081871} \\
\midrule
Mistral 7B & No   & \prop{0.1988}   & \prop{0.6082}   & \prop{0.0585}   & \prop{0.0585} \\
Mistral 7B & Last & \prop{0.163743} & \prop{0.625731} & \prop{0.076023} & \prop{0.076023} \\
Mistral 7B & Best & \prop{0.187135} & \prop{0.690058} & \prop{0.076023} & \prop{0.076023} \\
\midrule
Qwen 7B & No   & \prop{0.2632}   & \prop{0.1988}   & \prop{0.0234}   & \prop{0.0234} \\
Qwen 7B & Last & \prop{0.45614}  & \prop{0.064327} & \prop{0.035088} & \prop{0.023392} \\
Qwen 7B & Best & \prop{0.532164} & \prop{0.157895} & \prop{0.035088} & \prop{0.023392} \\
\midrule
QwQ 32B & No   & \prop{0.9883}   & \prop{0.8713}   & \prop{0.8713}   & \prop{0.807} \\
QwQ 32B & Last & \prop{1.0}      & \prop{0.988304} & \prop{0.988304} & \prop{0.918129} \\
QwQ 32B & Best & \prop{1.0}      & \prop{0.994152} & \prop{0.994152} & \prop{0.923977} \\
\bottomrule
\end{tabular}
\caption{Iterative prompting results with Basic prompting on an instance with 5-10 students. For each model, we have the metrics for no iterative prompting and iterative prompting with the last and best attempt, respectively.}
\label{iterative_prompting_by_model_basic}
\end{table}

\begin{table}[ht]
\begin{tabular}{llrrrr}
\toprule
\small
Model & Iterative prompting & Feasibility & Assignment & Matching & Optimality \\
\midrule
Llama 8B & Non   & \prop{0.4678}   & \prop{0.117}    & \prop{0.0409}  & \prop{0.0351}  \\
Llama 8B & Last & \prop{0.438596} & \prop{0.122807} & \prop{0.070175} & \prop{0.05848} \\
Llama 8B & Best & \prop{0.602339} & \prop{0.146199} & \prop{0.081871} & \prop{0.058480} \\
\midrule
Llama 70B & Non   & \prop{0.6316}   & \prop{0.0819}   & \prop{0.0468}   & \prop{0.0234} \\
Llama 70B & Last & \prop{0.853801} & \prop{0.116959} & \prop{0.111111} & \prop{0.081871} \\
Llama 70B & Best & \prop{0.888889} & \prop{0.140351} & \prop{0.140351} & \prop{0.081871} \\
\midrule
Mistral 7B& Non   & \prop{0.2242}   & \prop{0.5614}  & \prop{0.0585}  & \prop{0.0585} \\
Mistral 7B& Last & \prop{0.192982} & \prop{0.567251} & \prop{0.081871} & \prop{0.081871} \\
Mistral 7B& Best & \prop{0.228070} & \prop{0.614035} & \prop{0.081871} & \prop{0.081871} \\
\midrule
Qwen 7B& Non   & \prop{0.2339}   & \prop{0.2632}   & \prop{0.0351}  & \prop{0.0351} \\
Qwen 7B& Last & \prop{0.438596} & \prop{0.076023} & \prop{0.052632} & \prop{0.046784} \\
Qwen 7B& Best & \prop{0.549708} & \prop{0.19883}  & \prop{0.05848}  & \prop{0.046784} \\
\midrule
QwQ 32B& Non   & \prop{1}       & \prop{0.9181}   & \prop{0.9181}   & \prop{0.8304}   \\
QwQ 32B& Last & \prop{1.0}     & \prop{0.994152} & \prop{0.994152} & \prop{0.953216} \\
QwQ 32B& Best & \prop{1.0}     & \prop{0.988304} & \prop{0.988304} & \prop{0.94152}  \\

\bottomrule
\end{tabular}
\caption{Iterative prompting with Role prompting for instance with 5-10 students. For each model, we have the metrics for no iterative prompting and iterative prompting with the last and best attempt}
\label{iterative_prompting_by_model_role}
\end{table}

\begin{table}[ht]
\begin{tabular}{llrrrr}
\toprule
\small
Model & Iterative prompting & Feasibility & Assignment & Matching & Optimality \\
\midrule
Llama8B & No   & \prop{0.3333}   & \prop{0.1871}   & \prop{0.0585}   & \prop{0.0468} \\
Llama8B & Last & \prop{0.333333} & \prop{0.169591} & \prop{0.052632} & \prop{0.046784} \\
Llama8B & Best & \prop{0.298246} & \prop{0.175439} & \prop{0.05848}  & \prop{0.052632} \\

\midrule
Llama70B & No   & \prop{0.6257}   & \prop{0.1754}   & \prop{0.0819}   & \prop{0.0585} \\
Llama70B & Last & \prop{0.795322} & \prop{0.152047} & \prop{0.134503} & \prop{0.116959} \\
Llama70B & Best & \prop{0.853801} & \prop{0.192982} & \prop{0.157895} & \prop{0.116959} \\
\midrule
Mistral 7B & No   & \prop{0.2182}   & \prop{0.5758}   & \prop{0.0727}   & \prop{0.0727} \\
Mistral 7B & Last & \prop{0.192982} & \prop{0.590643} & \prop{0.081871} & \prop{0.081871} \\
Mistral 7B & Best & \prop{0.19883}  & \prop{0.596491} & \prop{0.081871} & \prop{0.081871} \\
\midrule
Qwen 7B & No   & \prop{0.3041}   & \prop{0.152}    & \prop{0.0409}   & \prop{0.0409} \\
Qwen 7B & Last & \prop{0.438596} & \prop{0.081871} & \prop{0.040936} & \prop{0.035088} \\
Qwen 7B & Best & \prop{0.467836} & \prop{0.128655} & \prop{0.05848}  & \prop{0.040936} \\
\midrule
QwQ 32B & No   & \prop{1.0}      & \prop{0.8824}   & \prop{0.8824}   & \prop{0.8118} \\
QwQ 32B & Last & \prop{1.0}      & \prop{0.976608} & \prop{0.976608} & \prop{0.929825} \\
QwQ 32B & Best & \prop{1.0}      & \prop{0.976608} & \prop{0.976608} & \prop{0.918129} \\
\bottomrule
\end{tabular}
\caption{Iterative prompting results with ICL prompting on an instance with 5-10 students. For each model, we have the metrics for no iterative prompting and iterative prompting with the last and best attempt, respectively.}
\label{iterative_prompting_by_model_icl}
\end{table}

\begin{table}[ht]
\begin{tabular}{llrrrr}
\toprule
\small
Model & Iterative prompting & Feasibility & Assignment & Matching & Optimality \\
\midrule
Llama8B & No   & \prop{0.5146}   & \prop{0.0409}   & \prop{0.0117}   & \prop{0.0117} \\
Llama8B & Last & \prop{0.368421} & \prop{0.093567} & \prop{0.046784} & \prop{0.02924} \\
Llama8B & Best & \prop{0.573099} & \prop{0.105263} & \prop{0.046784} & \prop{0.029240} \\

\midrule
Llama70B & No   & \prop{0.5205}   & \prop{0.2105}   & \prop{0.0877}   & \prop{0.0643} \\
Llama70B & Last & \prop{0.830409} & \prop{0.146199} & \prop{0.140351} & \prop{0.099415} \\
Llama70B & Best & \prop{0.859649} & \prop{0.187135} & \prop{0.152047} & \prop{0.099415} \\
\midrule
Mistral 7B & No   & \prop{0.1813}   & \prop{0.6959}   & \prop{0.0877}   & \prop{0.0819} \\
Mistral 7B & Last & \prop{0.210526} & \prop{0.608187} & \prop{0.099415} & \prop{0.093567} \\
Mistral 7B & Best & \prop{0.210526} & \prop{0.684211} & \prop{0.099415} & \prop{0.093567} \\
\midrule
Qwen 7B & No   & \prop{0.269}    & \prop{0.1754}   & \prop{0.0526}   & \prop{0.0526} \\
Qwen 7B & Last & \prop{0.339181} & \prop{0.093567} & \prop{0.064327} & \prop{0.052632} \\
Qwen 7B & Best & \prop{0.374269} & \prop{0.146199} & \prop{0.070175} & \prop{0.052632} \\
\midrule
QwQ 32B & No   & \prop{0.9883}   & \prop{0.8772}   & \prop{0.8772}   & \prop{0.8304} \\
QwQ 32B & Last & \prop{1.0}      & \prop{0.994152} & \prop{0.994152} & \prop{0.964912} \\
QwQ 32B & Best & \prop{1.0}      & \prop{1.0}      & \prop{1.0}      & \prop{0.97076} \\
\bottomrule
\end{tabular}
\caption{Iterative prompting results with ICL with steps prompting on an instance with 5-10 students. For each model, we have the metrics for no iterative prompting and iterative prompting with the last and best attempt, respectively.}
\label{iterative_prompting_by_model_icl_steps}
\end{table}

\begin{table}[ht]
\begin{tabular}{llrrrr}
\toprule
\small
Model & Iterative prompting & Feasibility & Assignment & Matching & Optimality \\
\midrule
Llama8B & No   & \prop{0.5029}   & \prop{0.0994}   & \prop{0.0292}   & \prop{0.0234} \\
Llama8B & Last & \prop{0.497076} & \prop{0.105263} & \prop{0.070175} & \prop{0.046784} \\
Llama8B & Best & \prop{0.660819} & \prop{0.134503} & \prop{0.076023} & \prop{0.046784} \\

\midrule
Llama70B & No   & \prop{0.655}    & \prop{0.076}    & \prop{0.0409}   & \prop{0.0234} \\
Llama70B & Last & \prop{0.894737} & \prop{0.122807} & \prop{0.122807} & \prop{0.070175} \\
Llama70B & Best & \prop{0.900585} & \prop{0.134503} & \prop{0.128655} & \prop{0.070175} \\
\midrule
Mistral 7B & No   & \prop{0.2222}   & \prop{0.6023}   & \prop{0.0526}   & \prop{0.0526} \\
Mistral 7B & Last & \prop{0.181287} & \prop{0.637427} & \prop{0.070175} & \prop{0.070175} \\
Mistral 7B & Best & \prop{0.216374} & \prop{0.684211} & \prop{0.070175} & \prop{0.070175} \\
\midrule
Qwen 7B & No   & \prop{0.4152}   & \prop{0.1345}   & \prop{0.0234}   & \prop{0.0234} \\
Qwen 7B & Last & \prop{0.479532} & \prop{0.064327} & \prop{0.02924}  & \prop{0.023392} \\
Qwen 7B & Best & \prop{0.625731} & \prop{0.105263} & \prop{0.029240} & \prop{0.023392} \\

\midrule
QwQ 32B & No   & \prop{1.0}      & \prop{0.9532}   & \prop{0.9532}   & \prop{0.8713} \\
QwQ 32B & Last & \prop{1.0}      & \prop{1.0}      & \prop{1.0}      & \prop{0.94152} \\
QwQ 32B & Best & \prop{1.0}      & \prop{1.0}      & \prop{1.0}      & \prop{0.929825} \\
\bottomrule
\end{tabular}
\caption{Iterative prompting results with CoT unsupervised prompting on an instance with 5-10 students. For each model, we have the metrics for no iterative prompting and iterative prompting with the last and best attempt, respectively.}
\label{iterative_prompting_by_model_unsup}
\end{table}

\begin{table}[ht]
\begin{tabular}{llrrrr}
\toprule
\small
Model & Iterative prompting & Feasibility & Assignment & Matching & Optimality \\
\midrule
Llama8B & No   & \prop{0.6608}   & \prop{0.0526}   & \prop{0.0117}   & \prop{0.0117} \\
Llama8B & Last & \prop{0.467836} & \prop{0.093567} & \prop{0.046784} & \prop{0.040936} \\
Llama8B & Best & \prop{0.760234} & \prop{0.116959} & \prop{0.052632} & \prop{0.040936} \\
\midrule
Llama70B & No   & \prop{0.6257}   & \prop{0.1345}   & \prop{0.0702}   & \prop{0.0468} \\
Llama70B & Last & \prop{0.812865} & \prop{0.134503} & \prop{0.128655} & \prop{0.099415} \\
Llama70B & Best & \prop{0.853801} & \prop{0.192982} & \prop{0.157895} & \prop{0.099415} \\
\midrule
Mistral 7B & No   & \prop{0.1988}   & \prop{0.6199}   & \prop{0.0643}   & \prop{0.0643} \\
Mistral 7B & Last & \prop{0.19883}  & \prop{0.608187} & \prop{0.070175} & \prop{0.070175} \\
Mistral 7B & Best & \prop{0.204678} & \prop{0.637427} & \prop{0.070175} & \prop{0.070175} \\
\midrule
Qwen 7B & No   & \prop{0.2515}   & \prop{0.2573}   & \prop{0.0409}   & \prop{0.0409} \\
Qwen 7B & Last & \prop{0.356725} & \prop{0.122807} & \prop{0.046784} & \prop{0.046784} \\
Qwen 7B & Best & \prop{0.421053} & \prop{0.216374} & \prop{0.046784} & \prop{0.046784} \\
\midrule
QwQ 32B & No   & \prop{1.0}      & \prop{0.9649}   & \prop{0.9649}   & \prop{0.8947} \\
QwQ 32B & Last & \prop{1.0}      & \prop{0.994152} & \prop{0.994152} & \prop{0.947368} \\
QwQ 32B & Best & \prop{1.0}      & \prop{0.988304} & \prop{0.988304} & \prop{0.941520} \\
\bottomrule
\end{tabular}
\caption{Iterative prompting results with CoT text prompting on an instance with 5-10 students. For each model, we have the metrics for no iterative prompting and iterative prompting with the last and best attempt, respectively.}
\label{iterative_prompting_by_model_txt}
\end{table}

\begin{table}[ht]
\begin{tabular}{llrrrr}
\toprule
\small
Model & Iterative prompting & Feasibility & Assignment & Matching & Optimality \\
\midrule
Llama8B & No   & \prop{0.4912}   & \prop{0.117}    & \prop{0.0292}   & \prop{0.0292} \\
Llama8B & Last & \prop{0.461988} & \prop{0.111111} & \prop{0.064327} & \prop{0.064327} \\
Llama8B & Best & \prop{0.631579} & \prop{0.152047} & \prop{0.070175} & \prop{0.064327} \\
\midrule
Llama70B & No   & \prop{0.6023}   & \prop{0.1345}   & \prop{0.0585}   & \prop{0.0409} \\
Llama70B & Last & \prop{0.807018} & \prop{0.140351} & \prop{0.134503} & \prop{0.099415} \\
Llama70B & Best & \prop{0.830409} & \prop{0.210526} & \prop{0.157895} & \prop{0.099415} \\
\midrule
Mistral 7B & No   & \prop{0.1696}   & \prop{0.8129}   & \prop{0.0819}   & \prop{0.0819} \\
Mistral 7B & Last & \prop{0.187135} & \prop{0.736842} & \prop{0.087719} & \prop{0.087719} \\
Mistral 7B & Best & \prop{0.192982} & \prop{0.801170} & \prop{0.087719} & \prop{0.087719} \\
\midrule
Qwen 7B & No   & \prop{0.4211}   & \prop{0.1228}   & \prop{0.0292}   & \prop{0.0058} \\
Qwen 7B & Last & \prop{0.415205} & \prop{0.122807} & \prop{0.011696} & \prop{0.011696} \\
Qwen 7B & Best & \prop{0.461988} & \prop{0.152047} & \prop{0.035088} & \prop{0.011696} \\
\midrule
QwQ 32B & No   & \prop{0.9883}   & \prop{0.8947}   & \prop{0.8947}   & \prop{0.848} \\
QwQ 32B & Last & \prop{1.0}      & \prop{0.994152} & \prop{0.994152} & \prop{0.953216} \\
QwQ 32B & Best & \prop{1.0}      & \prop{0.988304} & \prop{0.988304} & \prop{0.947368} \\
\bottomrule
\end{tabular}
\caption{Iterative prompting results with CoT python prompting on an instance with 5-10 students. For each model, we have the metrics for no iterative prompting and iterative prompting with the last and best attempt, respectively.}
\label{iterative_prompting_by_model_python}
\end{table}

\begin{table}[ht]
\begin{tabular}{llrrrr}
\toprule
\small
Model & Iterative prompting & Feasibility & Assignment & Matching & Optimality \\
\midrule
Llama8B & No   & \prop{0.5205}   & \prop{0.076}    & \prop{0.0175}   & \prop{0.0175} \\
Llama8B & Last & \prop{0.45614}  & \prop{0.111111} & \prop{0.076023} & \prop{0.076023} \\
Llama8B & Best & \prop{0.625731} & \prop{0.163743} & \prop{0.087719} & \prop{0.076023} \\

\midrule
Llama70B & No   & \prop{0.6433}   & \prop{0.1637}   & \prop{0.0819}   & \prop{0.0585} \\
Llama70B & Last & \prop{0.748538} & \prop{0.157895} & \prop{0.152047} & \prop{0.122807} \\
Llama70B & Best & \prop{0.830409} & \prop{0.210526} & \prop{0.175439} & \prop{0.122807} \\
\midrule
Mistral 7B & No   & \prop{0.1754}   & \prop{0.7661}   & \prop{0.0877}   & \prop{0.0877} \\
Mistral 7B & Last & \prop{0.192982} & \prop{0.625731} & \prop{0.087719} & \prop{0.087719} \\
Mistral 7B & Best & \prop{0.192982} & \prop{0.754386} & \prop{0.087719} & \prop{0.087719} \\
\midrule
Qwen 7B & No   & \prop{0.3216}   & \prop{0.2281}   & \prop{0.0526}   & \prop{0.0468} \\
Qwen 7B & Last & \prop{0.397661} & \prop{0.099415} & \prop{0.064327} & \prop{0.05848} \\
Qwen 7B & Best & \prop{0.461988} & \prop{0.198830} & \prop{0.070175} & \prop{0.058480} \\
\midrule
QwQ 32B & No   & \prop{0.9883}   & \prop{0.9064}   & \prop{0.9006}   & \prop{0.8538} \\
QwQ 32B & Last & \prop{1.0}      & \prop{0.994152} & \prop{0.994152} & \prop{0.976608} \\
QwQ 32B & Best & \prop{1.0}      & \prop{0.994152} & \prop{0.994152} & \prop{0.976608} \\
\bottomrule
\end{tabular}
\caption{Iterative prompting results with CoT pseudocode prompting on an instance with 5-10 students. For each model, we have the metrics for no iterative prompting and iterative prompting with the last and best attempt, respectively.}
\label{iterative_prompting_by_model_pseudo}
\end{table}

\subsection{Graphs}
\begin{figure*}[]
  \centering

  \begin{subfigure}{0.45\textwidth}
    \includegraphics[width=\linewidth]{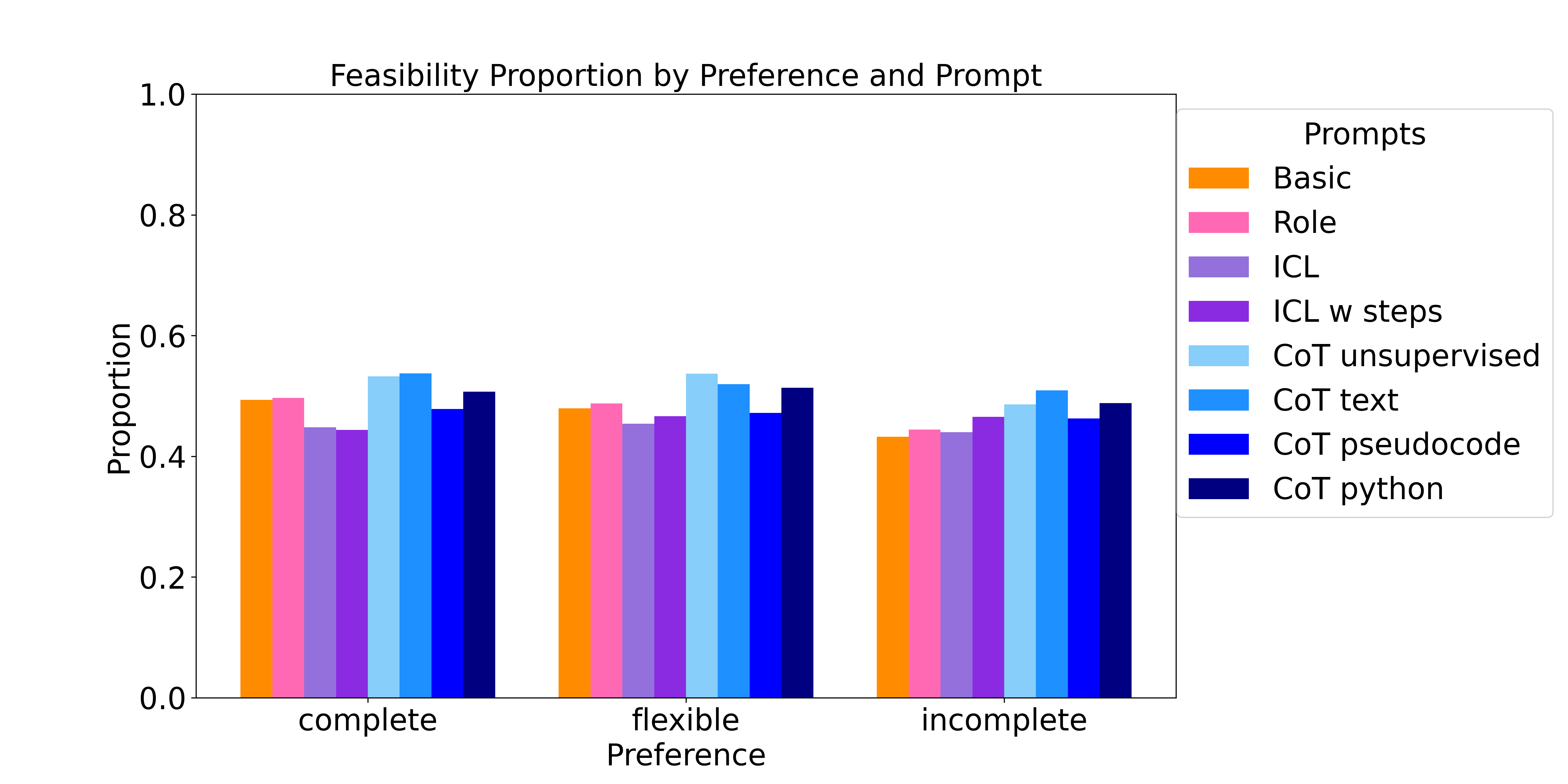}
    \caption{Feasibility}
    \label{fig:feasible_pref_prompt}
  \end{subfigure}
  \hfill
  \begin{subfigure}{0.45\textwidth}
    \includegraphics[width=\linewidth]{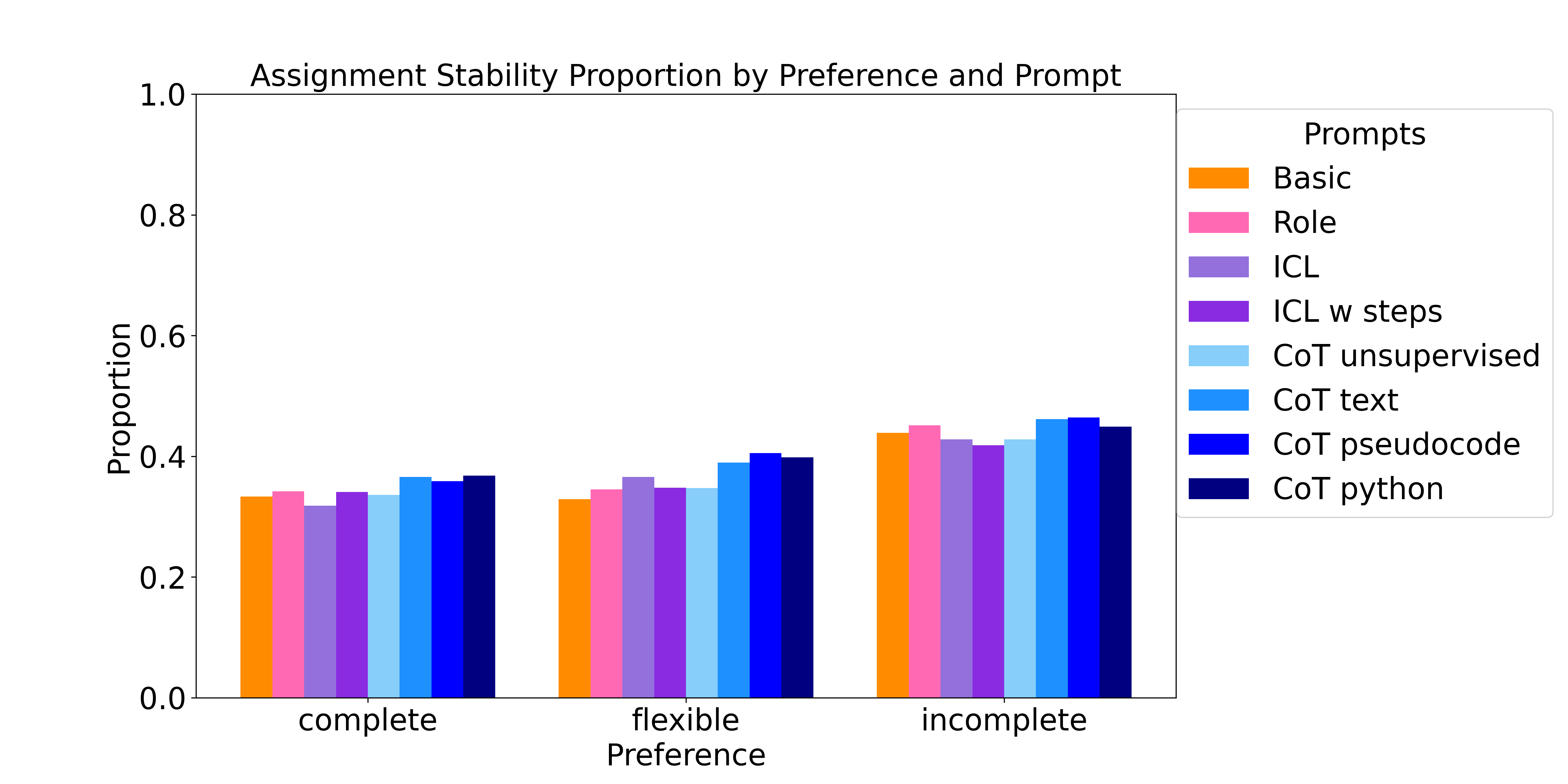}
    \caption{Assignment Stability}
    \label{fig:assignment_pref_prompt}
  \end{subfigure}

  \begin{subfigure}{0.45\textwidth}
    \includegraphics[width=\linewidth]{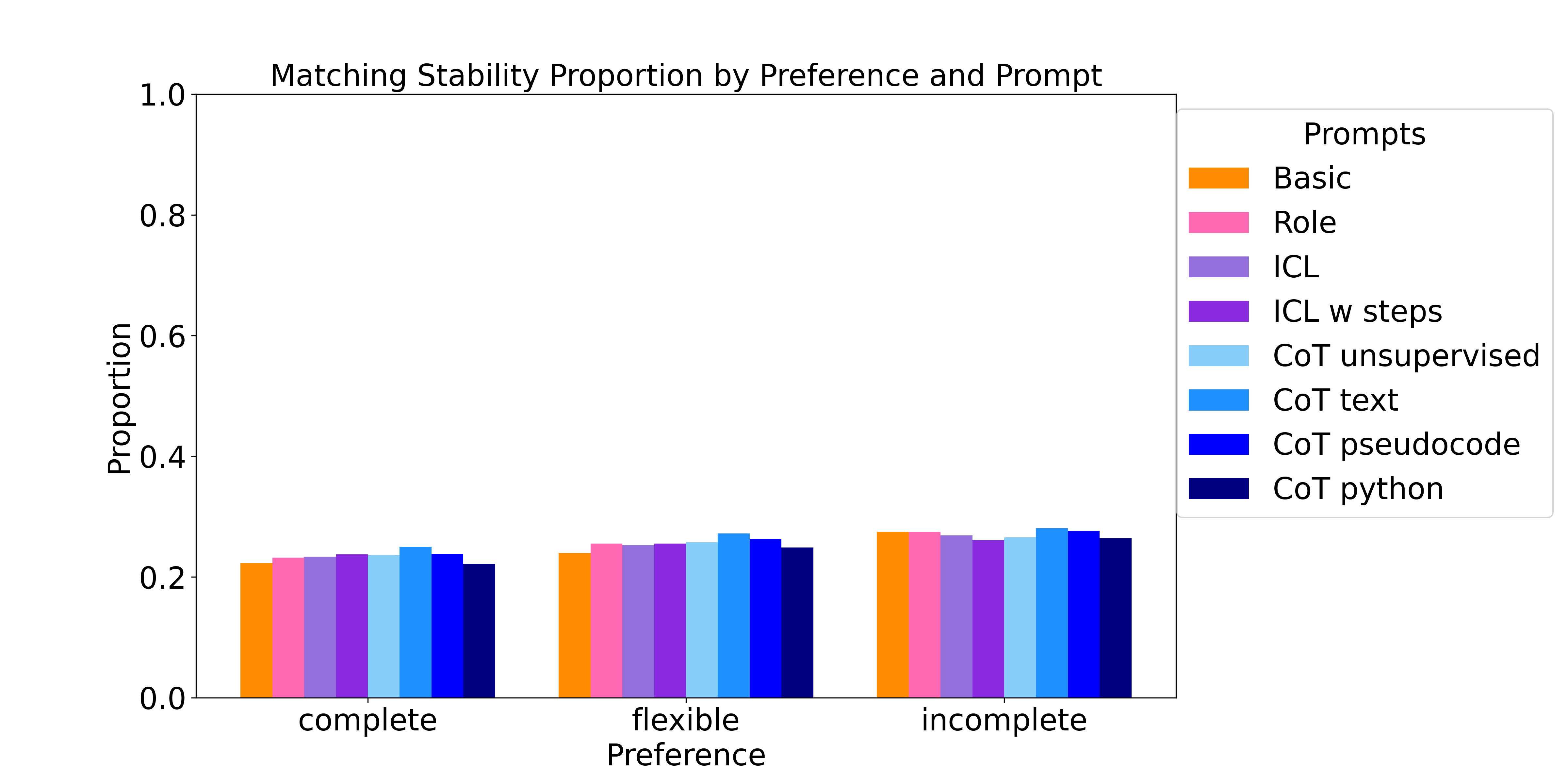}
    \caption{Matching Stability}
    \label{fig:matching_pref_prompt}
  \end{subfigure}
  \hfill
  \begin{subfigure}{0.45\textwidth}
    \includegraphics[width=\linewidth]{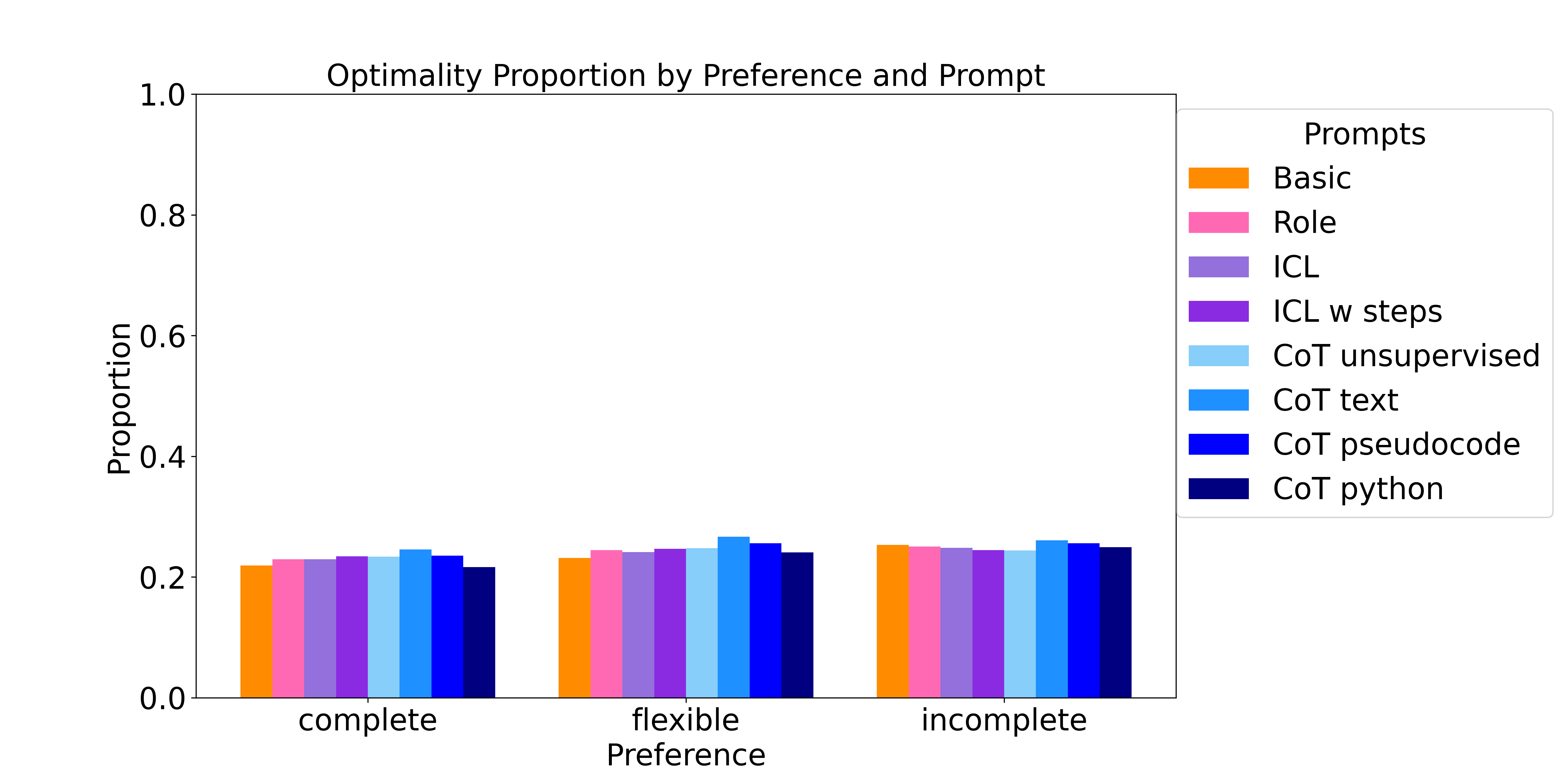}
    \caption{Optimality}
    \label{fig:optimal_pref_prompt}
  \end{subfigure}

  \caption{Proportion metrics by preference for all prompts.}
  \label{fig:pref_metrics_prompt}
\end{figure*}

\begin{figure*}[]
  \centering

  \begin{subfigure}{0.45\textwidth}
    \includegraphics[width=\linewidth]{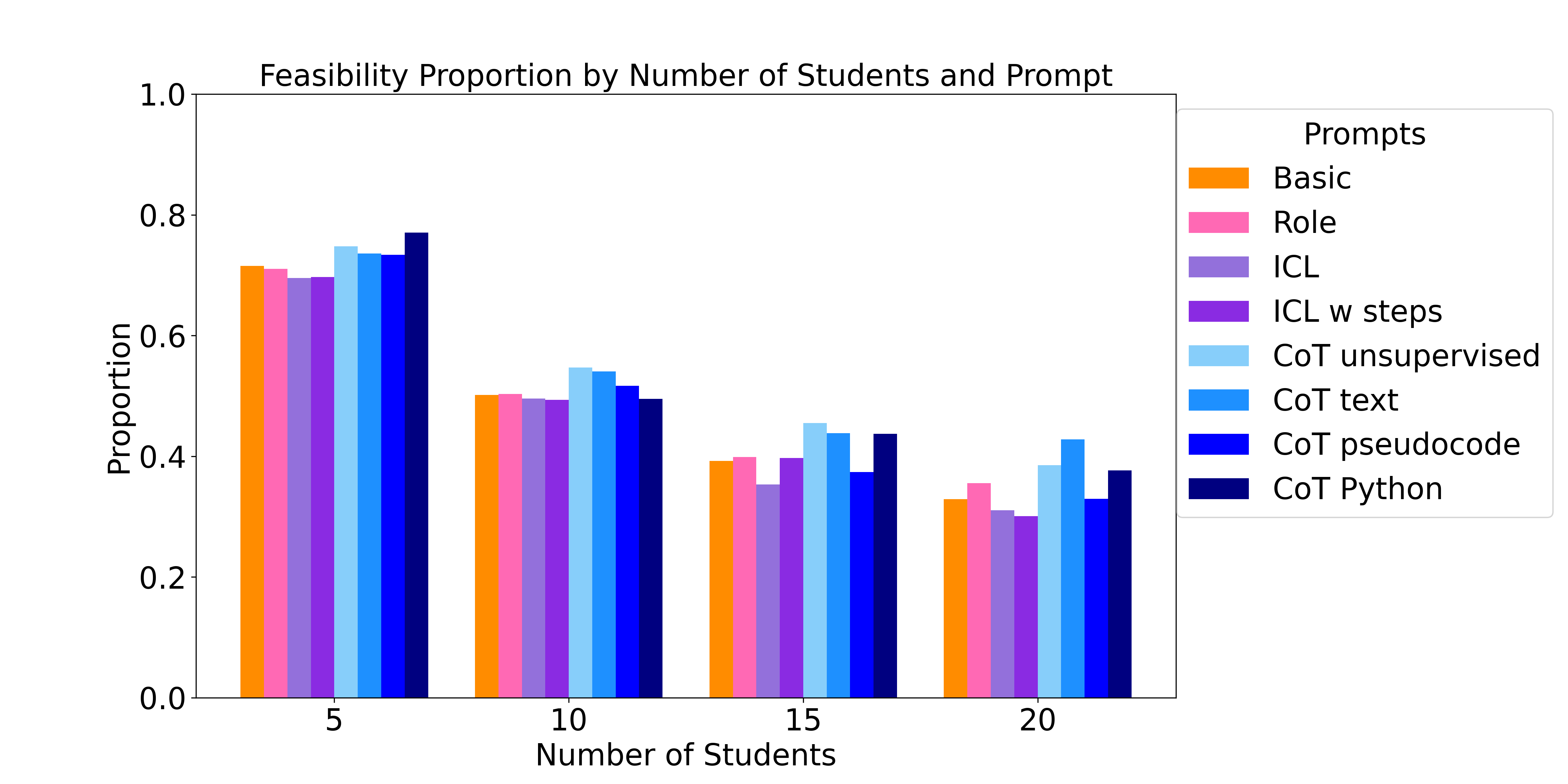}
    \caption{Feasibility}
    \label{fig:feasible_student_prompt}
  \end{subfigure}
  \hfill
  \begin{subfigure}{0.45\textwidth}
    \includegraphics[width=\linewidth]{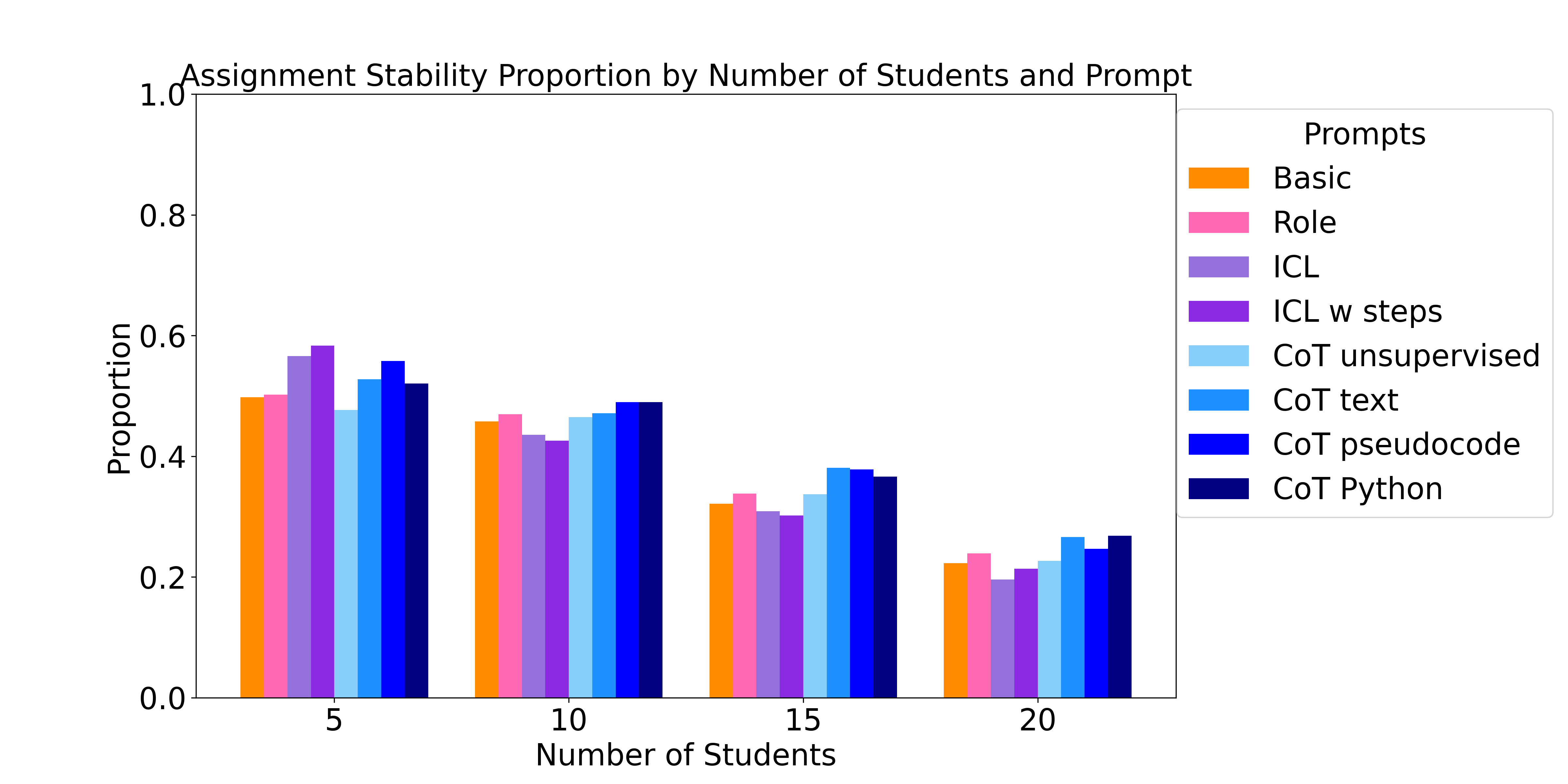}
    \caption{Assignment Stability}
    \label{fig:assignment_student_prompt}
  \end{subfigure}

  \begin{subfigure}{0.45\textwidth}
    \includegraphics[width=\linewidth]{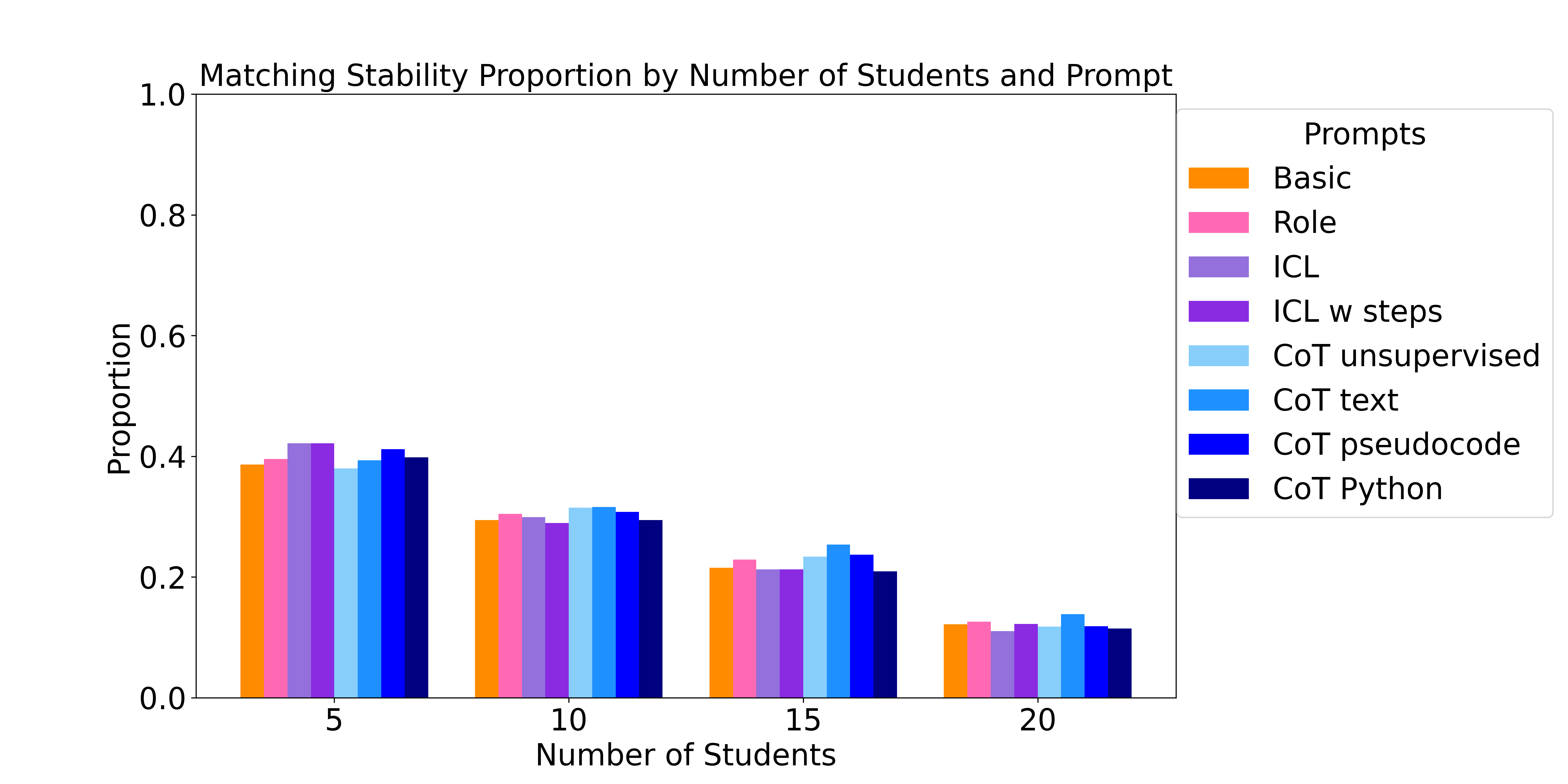}
    \caption{Matching Stability}
    \label{fig:matching_students_prompt}
  \end{subfigure}
  \hfill
  \begin{subfigure}{0.45\textwidth}
    \includegraphics[width=\linewidth]{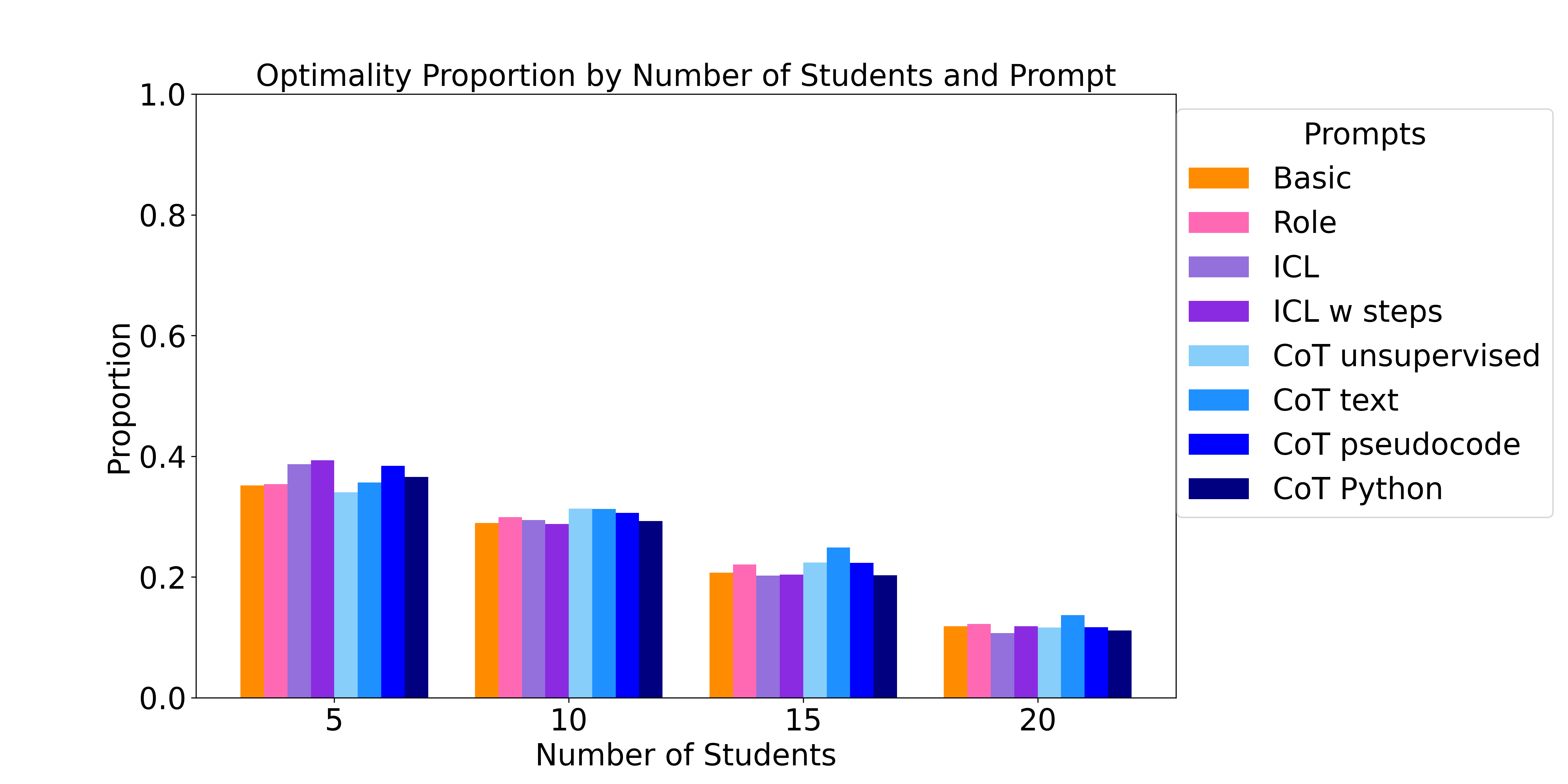}
    \caption{Optimality}
    \label{fig:optimal_students_prompt}
  \end{subfigure}

  \caption{Proportion metrics by number of students for all prompts.}
  \label{fig:students_metrics_prompt}
\end{figure*}

While the Figure~\ref{fig:spider_prmpt} presented the trend for every prompt, the Figure~\ref{fig:pref_metrics_prompt} and Figure~\ref{fig:students_metrics_prompt} show the detailed results for every prompt, under both complexity dimensions, the number of students and the preference types.

\end{document}